\documentclass[11pt]{article}

\usepackage[preprint]{acl}

\usepackage{times}
\usepackage{latexsym}

\usepackage[utf8]{inputenc}

\usepackage{microtype}

\usepackage{amsmath,amsfonts,bm}

\def\eqref#1{equation~\ref{#1}}

\def\1{\bm{1}}

\DeclareMathAlphabet{\mathsfit}{\encodingdefault}{\sfdefault}{m}{sl}
\SetMathAlphabet{\mathsfit}{bold}{\encodingdefault}{\sfdefault}{bx}{n}

\definecolor{mydarkblue}{rgb}{0,0.08,0.45}
\hypersetup{
    colorlinks = true,
    linkcolor = mydarkblue,
    urlcolor = mydarkblue,
    citecolor = mydarkblue
}
\hypersetup{pdftitle={Demystifying Hybrid Thinking: Can LLMs Truly Switch Between Think and No-Think?},pdfauthor={Shouren Wang, Wang Yang, Xianxuan Long, Debargha Ganguly, Qifan Wang, Vipin Chaudhary, Xiaotian Han}}
\usepackage{inconsolata}
\usepackage{url}
\usepackage[T1]{fontenc}

\usepackage{indentfirst}
\usepackage{xspace}
\usepackage{tabularx}
\usepackage{multirow}
\usepackage{adjustbox}
\usepackage[capitalize,noabbrev,nameinlink]{cleveref}
\usepackage{xcolor}
\usepackage{array} 
\usepackage{booktabs}
\usepackage{colortbl}
\usepackage{enumitem}
\usepackage{threeparttable}
\usepackage{tablefootnote}
\usepackage{graphicx}
\usepackage{subcaption}
\usepackage{enumitem}

\definecolor{darkgreen}{rgb}{0.0, 0.5, 0.0} 
\definecolor{lowblue}{rgb}{0.1,0.2,0.5} %

\title{Demystifying Hybrid Thinking: \\ Can LLMs Truly Switch Between Think and No-Think?}

\author{\noindent
\textbf{Shouren Wang$^{1, *}$,~Wang Yang$^{1, *}$,~Xianxuan Long$^{2}$,~Debargha Ganguly$^{1}$,~Qifan Wang$^{3}$}\\
\textbf{Vipin Chaudhary$^{1, \dagger}$, Xiaotian Han$^{1, \dagger}$}\\
$^{1}$Case Western Reserve University \\
\texttt{\{sxw992,wxy320,dxg512,vipin,xhan\}@case.edu} \\
$^{2}$Michigan State University \quad $^{3}$Meta AI \\
\texttt{longxia2@msu.edu} \quad \texttt{wqfcr@meta.com} \\
$^{*}$Equal contribution \quad $^{\dagger}$Corresponding authors
}

\newcommand{\nothink}{\texttt{\textbackslash no\_think}}

\begin{document}

\maketitle

\begin{abstract}

Hybrid thinking enables LLMs to switch between reasoning and direct answering, offering a balance between efficiency and reasoning capability. Yet our experiments reveal that current hybrid thinking LLMs only achieve partial mode separation: reasoning behaviors often leak into the no-think mode. To understand and mitigate this, we analyze the factors influencing controllability and identify four that matter most: (1) larger data scale, (2) using think and no-think answers from different questions rather than the same question, (3) a moderate increase in no-think data number, and (4) a two-phase strategy that first trains reasoning ability and then applies hybrid think training. Building on these findings, we propose a practical recipe that, compared to standard training, can maintain accuracy in both modes while significantly reducing no-think output length (from $1085$ to $585$ on MATH500) and occurrences of reasoning-supportive tokens such as ``\texttt{wait}'' (from $5917$ to $522$ on MATH500). Our findings highlight the limitations of current hybrid thinking and offer directions for strengthening its controllability. The code is available at: \url{https://github.com/SR-A-W/demystifying-hybrid-thinking}.

\end{abstract}

\section{Introduction}
\begin{figure}[ht]
\centering
\includegraphics[width=\linewidth]{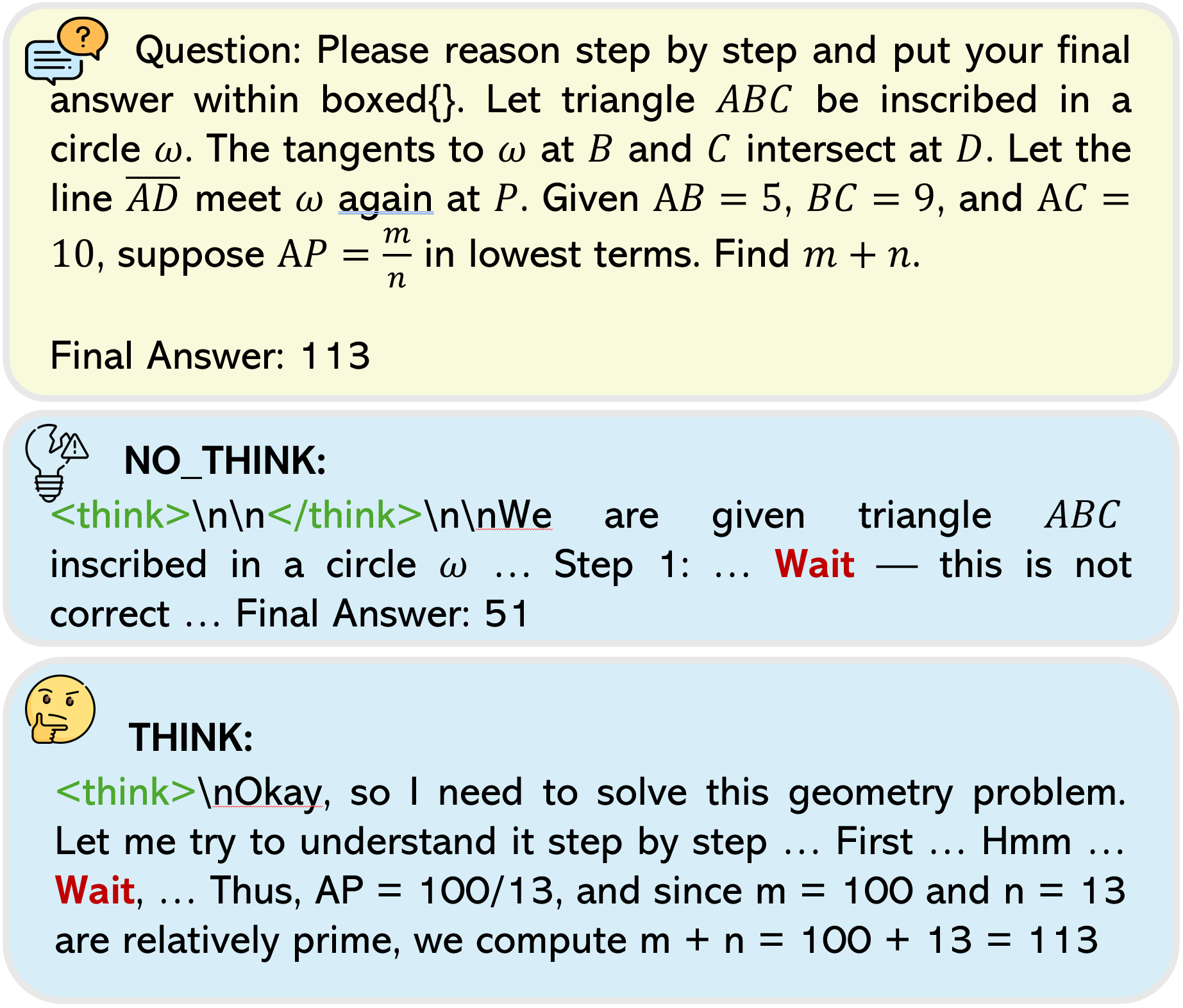}
\vspace{-15pt}
\caption{Case study on an AIME24 problem. We compare the responses of Qwen3-8B under no-think and think modes. In the no-think mode, Qwen3-8B still performs reasoning outside the no-think constraint (e.g., generating reasoning-supportive words such as \textcolor{red}{\texttt{Wait}}), indicating that its hybrid thinking ability remains imperfect and cannot achieve full control.}
\label{fig:motivation}
\end{figure}

Hybrid thinking is a widely adopted approach in many company models, enabling control over whether the model engages in reasoning and thereby achieving a more efficient and flexible reasoning process. Examples include Gemini~\citep{comanici2025gemini}, GPT-oss~\citep{agarwal2025gpt}, Qwen3~\citep{yang2025qwen3}, and DeepSeek V3.1~\citep{deepseekai2025v31}. However, few studies have systematically investigated the capabilities of hybrid thinking models, such as the training factors that influence them and the limitations they face in practice. This paper fills this gap by 1) offering a comprehensive analysis of hybrid thinking and 2) introducing a practical training recipe to enhance its controllability.

\begin{table*}[h!]
\centering
\caption{Performance on MATH500, AIME24, and GPQA under think and no-think modes. 
We report accuracy (Acc.), output length (Len.), \#Wait count (i.e., occurrences of reflection words such as ``\texttt{wait}'', ``\texttt{hmm}'', ``\texttt{alternatively}'', aggregated over all responses in the evaluation), and \textbf{no-think differences} ($\Delta$) between Hybrid (Qwen3-8B) and Instruct (Qwen2.5-7B-Instruct). 
Notably, Hybrid model in the no-think mode produce much longer outputs than the Instruct model, and even generate reflection words.}
\label{tab:mov_all}
\vspace{-8pt}
\setlength{\tabcolsep}{4pt}
\begin{tabular}{l|ccc|ccc|ccc}
\toprule
\multirow{2}{*}{Model} 
& \multicolumn{3}{c|}{Acc. (\%)} & \multicolumn{3}{c|}{Len.} & \multicolumn{3}{c}{\#Wait} \\
\cmidrule(lr){2-4} \cmidrule(lr){5-7} \cmidrule(lr){8-10}
& Think & No-think & $\Delta$ & Think & No-think & $\Delta$ & Think & No-think & $\Delta$ \\
\midrule
\multicolumn{10}{c}{MATH500} \\
\midrule
Instruct & --    & \textcolor{lowblue}{59.94} &  & --    & \textcolor{lowblue}{703.11}  &  & -- & \textcolor{lowblue}{0}   &  \\
Hybrid   & 92.82 & \textcolor{lowblue}{82.90} & \textcolor{lowblue}{\textbf{+22.96}} & 4384.35 & \textcolor{lowblue}{958.31} & \textcolor{lowblue}{\textbf{+255.20}} & 83703 & \textcolor{lowblue}{646} & \textcolor{lowblue}{\textbf{+646}} \\
\midrule
\multicolumn{10}{c}{AIME24} \\
\midrule
Instruct & --    & \textcolor{lowblue}{6.67}  &  & --    & \textcolor{lowblue}{1729.22} &  & -- & \textcolor{lowblue}{0}   &  \\
Hybrid   & 63.33 & \textcolor{lowblue}{24.00} & \textcolor{lowblue}{\textbf{+17.33}} & 11394.54 & \textcolor{lowblue}{4061.67} & \textcolor{lowblue}{\textbf{+2332.45}} & 12184 & \textcolor{lowblue}{184} & \textcolor{lowblue}{\textbf{+184}} \\
\midrule
\multicolumn{10}{c}{GPQA} \\
\midrule
Instruct & --    & \textcolor{lowblue}{30.15} &  & --    & \textcolor{lowblue}{775.07}  &  & -- & \textcolor{lowblue}{0}   &  \\
Hybrid   & 59.14 & \textcolor{lowblue}{47.93} & \textcolor{lowblue}{\textbf{+17.78}} & 7451.07 & \textcolor{lowblue}{1364.96} & \textcolor{lowblue}{\textbf{+589.89}} & 66152 & \textcolor{lowblue}{616} & \textcolor{lowblue}{\textbf{+616}} \\
\bottomrule
\end{tabular}
\end{table*}

We begin by exploring the output differences of hybrid thinking models. Hybrid thinking is typically implemented by adding control tokens (e.g., \texttt{\textbackslash no\_think}, \texttt{\textbackslash think}) into the prompt and applying supervised fine-tuning (SFT). Evaluating Qwen3-8B on MATH500, AIME24 and GPQA, with Qwen2.5-7B-Instruct as a pure no-think baseline, we find that hybrid models perform better in the think mode than in the no-think mode (e.g. 63\% accuracy and 11,394 tokens in
think vs. 24\% and 4,062 in no-think), suggesting that they do learn to use control tokens to modulate outputs.  

However, comparison with Qwen2.5-7B-Instruct shows that thinking mode switch is incomplete: Qwen2.5-7B-Instruct generates short outputs without reasoning. But, in the ``\textit{no\_think}'' mode, Qwen3-8B still generate outputs with reasoning, which leaks reasoning words (e.g., ``\texttt{wait}'', ``\texttt{hmm}'') outside empty ``\texttt{<think>} '' blocks (\Cref{fig:motivation}). Therefore, hybrid thinking affords only limited control, motivating further study of training strategies and their trade-offs.

To investigate which factors affect the controllability of hybrid thinking---i.e., the balance between think and no-think mode---we systematically analyze and obtain the following findings: 
\begin{enumerate}
    \item \textit{Large-scale data enables effective hybrid switching.} A sufficiently large amount of hybrid thinking data (e.g., 140k samples) is required for stable control.  
    \item \textit{No-think control improves with non-paired data.} Using think and no-think answers from different questions (\textit{no-pairs}) yields stronger controllability than paired settings (think and no-think answers from the same questions).  
    \item \textit{A moderate rise in no-think data strengthens control.} Appropriately increasing the no-think data proportion reduces output length in the no-think mode while maintaining accuracy.  
    \item \textit{Two-phase training enhances no-think control.} First training on pure think data, then applying hybrid training, further improves the controllability of no-think mode.  
\end{enumerate}

\begin{figure*}[t]
\centering
\begin{subfigure}[t]{0.32\linewidth}
    \centering
    \includegraphics[width=\linewidth]{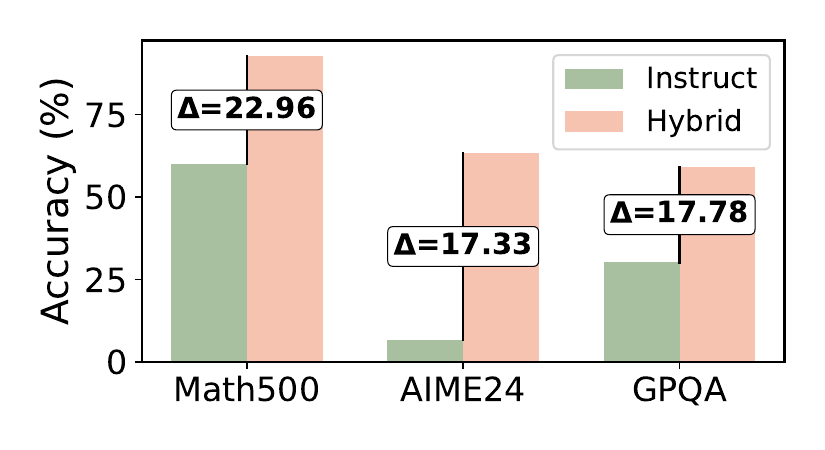}
    \caption{Acc}
\end{subfigure}
\hfill
\begin{subfigure}[t]{0.32\linewidth}
    \centering
    \includegraphics[width=\linewidth]{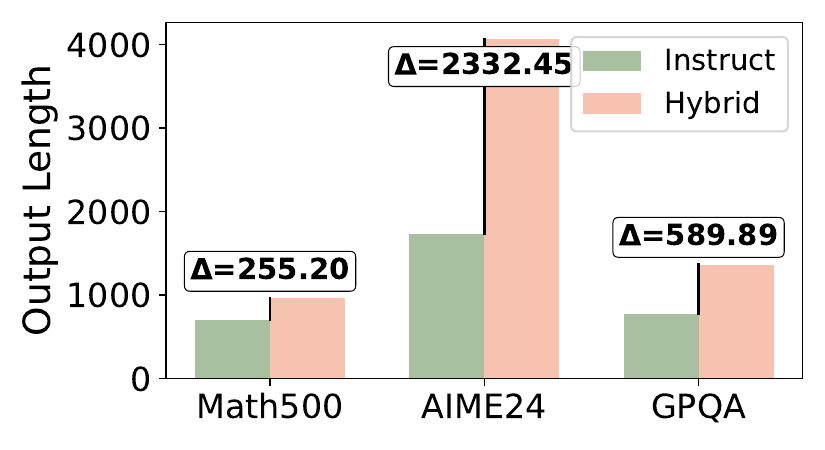}
    \caption{Length}
\end{subfigure}
\hfill
\begin{subfigure}[t]{0.32\linewidth}
    \centering
    \includegraphics[width=\linewidth]{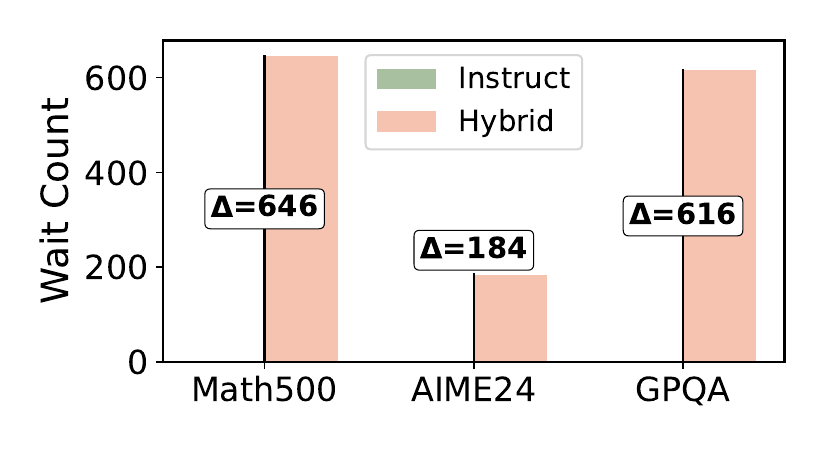}
    \caption{Wait Count}
\end{subfigure}
\vspace{-6pt}
\caption{Comparison of the performance of \textbf{no-think} mode of Hybrid model (Qwen3-8B) and Instruct model (Qwen2.5-7B-Instruct)   on MATH500, AIME24, and GPQA. We show bar plots of accuracy, average output length and wait count across these three datasets.}\label{fig:mov_all}
\end{figure*}

\begin{table*}[t]
\centering
\caption{Model-based judge validation of \#Refl./Ans.\ as a genuine (not surface-level) leakage signal, on off-the-shelf Qwen3 hybrid baselines. Judge: GPT-OSS-20B, three-way no/implicit/explicit verdict.}
\label{tab:appendix_judge_validation}
\resizebox{\textwidth}{!}{
\begin{tabular}{llcccc}
\toprule
Benchmark & Model & Mode & Output Len. & \#Refl. & Judge verdict (agreement) \\
\midrule
\multirow{6}{*}{MATH500}
 & Qwen3-4B  & No-think & 969  & 210    & Implicit reasoning (100\%) \\
 & Qwen3-4B  & Think    & 4750 & 163160 & Explicit reasoning (98\%)  \\
 & Qwen3-8B  & No-think & 1044 & 113    & Implicit reasoning (92\%)  \\
 & Qwen3-8B  & Think    & 4419 & 154670 & Explicit reasoning (98\%)  \\
 & Qwen3-14B & No-think & 885  & 131    & Implicit reasoning (98\%)  \\
 & Qwen3-14B & Think    & 4468 & 127342 & Explicit reasoning (98\%)  \\
\midrule
\multirow{6}{*}{AIME24}
 & Qwen3-4B  & No-think & 5022  & 672   & Implicit reasoning (94\%)  \\
 & Qwen3-4B  & Think    & 11410 & 21481 & Explicit reasoning (94\%)  \\
 & Qwen3-8B  & No-think & 4677  & 24    & Implicit reasoning (98\%)  \\
 & Qwen3-8B  & Think    & 11578 & 23222 & Explicit reasoning (96\%)  \\
 & Qwen3-14B & No-think & 4013  & 74    & Implicit reasoning (98\%)  \\
 & Qwen3-14B & Think    & 11341 & 18814 & Explicit reasoning (98\%)  \\
\midrule
\multirow{6}{*}{GPQA-Diamond}
 & Qwen3-4B  & No-think & 1504 & 150    & Implicit reasoning (100\%) \\
 & Qwen3-4B  & Think    & 7615 & 132680 & Explicit reasoning (100\%) \\
 & Qwen3-8B  & No-think & 1245 & 196    & Implicit reasoning (96\%)  \\
 & Qwen3-8B  & Think    & 7481 & 131701 & Explicit reasoning (98\%)  \\
 & Qwen3-14B & No-think & 1294 & 181    & Implicit reasoning (98\%)  \\
 & Qwen3-14B & Think    & 7246 & 105493 & Explicit reasoning (96\%)  \\
\bottomrule
\end{tabular}
}
\end{table*}
Finally, we compare hybrid thinking models with those trained purely on think or no-think data, under carefully matched training settings. Our results show that hybrid thinking cannot achieve complete mode separation, as the no-think mode remains influenced by think data. Based on our analysis of training factors, we propose a practical recipe to improve controllability in hybrid thinking, which maintains accuracy in both two modes while substantially reducing no-think verbosity. For example, on MATH500, the average no-think output length decreases from 1085 to 585 tokens, and the number of ``\texttt{wait}'' occurrences is reduced from 5917 to 522. These results demonstrate that our recipe effectively improves controllability without sacrificing performance, highlighting both the limitations of current hybrid thinking and concrete directions for its advancement.

We thus advocate that future hybrid thinking training should deliberately allocate appropriately more no-think data and adopt structured strategies such as two-phase training, which we believe will be crucial for advancing the development of controllable and efficient reasoning models.

\begin{figure*}[t]
\centering
\begin{subfigure}[t]{0.24\linewidth}
    \centering
    \includegraphics[width=\linewidth]{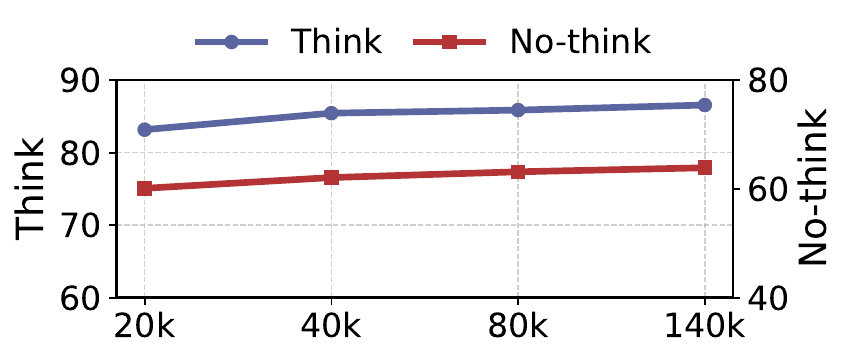}
    \caption{Acc on MATH500}
\end{subfigure}
\hfill
\begin{subfigure}[t]{0.24\linewidth}
    \centering
    \includegraphics[width=\linewidth]{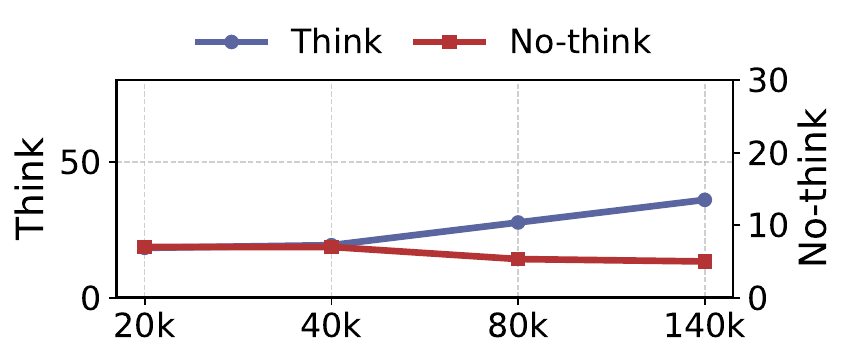}
    \caption{Acc on AIME24}
\end{subfigure}
\hfill
\begin{subfigure}[t]{0.24\linewidth}
    \centering
    \includegraphics[width=\linewidth]{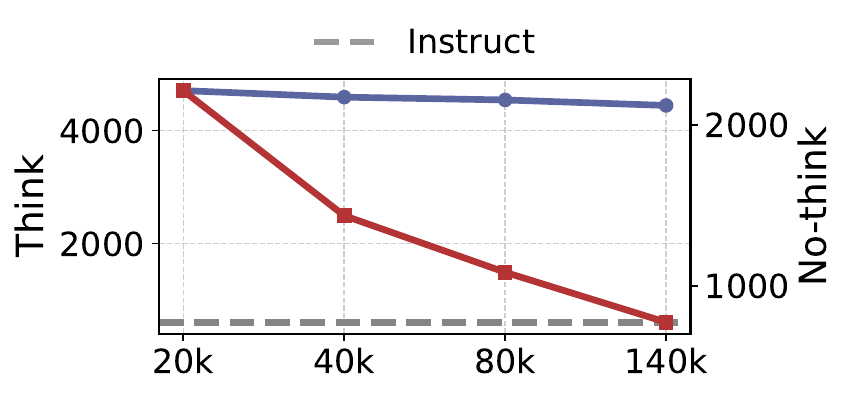}
    \caption{Len. on MATH500}
    \label{fig:math500_length}
\end{subfigure}
\hfill
\begin{subfigure}[t]{0.24\linewidth}
    \centering
    \includegraphics[width=\linewidth]{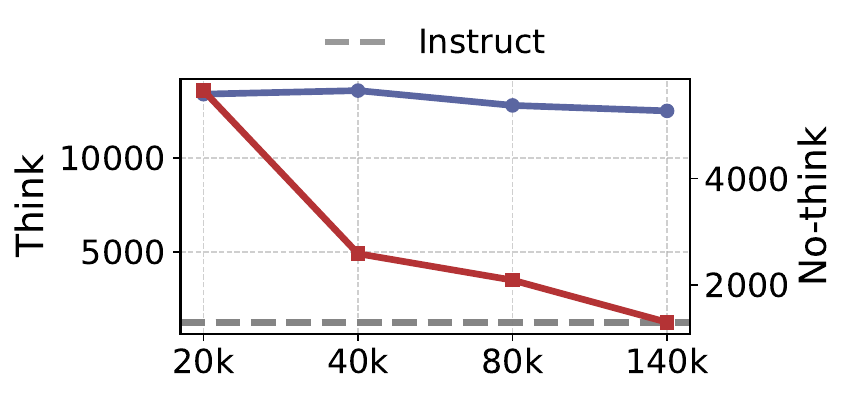}
    \caption{Len. on AIME24}
    \label{fig:aime24_length}
\end{subfigure}
\vspace{-6pt}
\caption{Line plots of Qwen2.5-7B-Instruct performance with different training data scales (20k/40k/80k/140k) on MATH500, AIME24, and GPQA. We report accuracy (top) and average output length (bottom) under both think and no-think modes. The results show that while accuracy and output length in the think mode, as well as accuracy in the no-think mode, remain relatively stable across scales, the no-think output length decreases substantially with larger training data scales as expected.}
\label{fig:line plot data scale}
\end{figure*}

\begin{figure*}[t]
\centering
\begin{subfigure}[t]{0.24\linewidth}
    \centering
    \includegraphics[width=\linewidth]{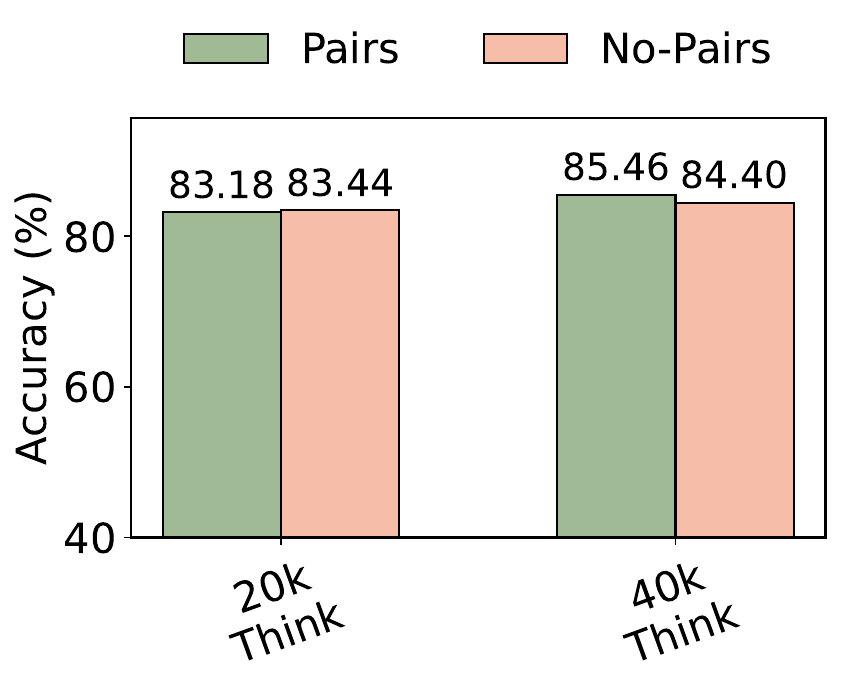}
    \caption{Think Acc.}
\end{subfigure}
\hfill
\begin{subfigure}[t]{0.24\linewidth}
    \centering
    \includegraphics[width=\linewidth]{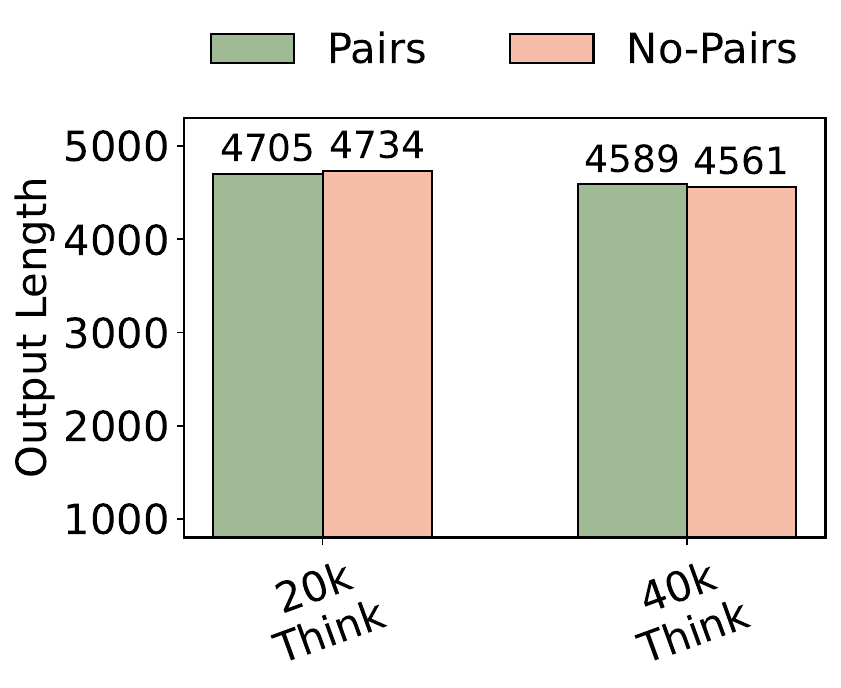}
    \caption{Think Len.}
\end{subfigure}
\hfill
\begin{subfigure}[t]{0.24\linewidth}
    \centering
    \includegraphics[width=\linewidth]{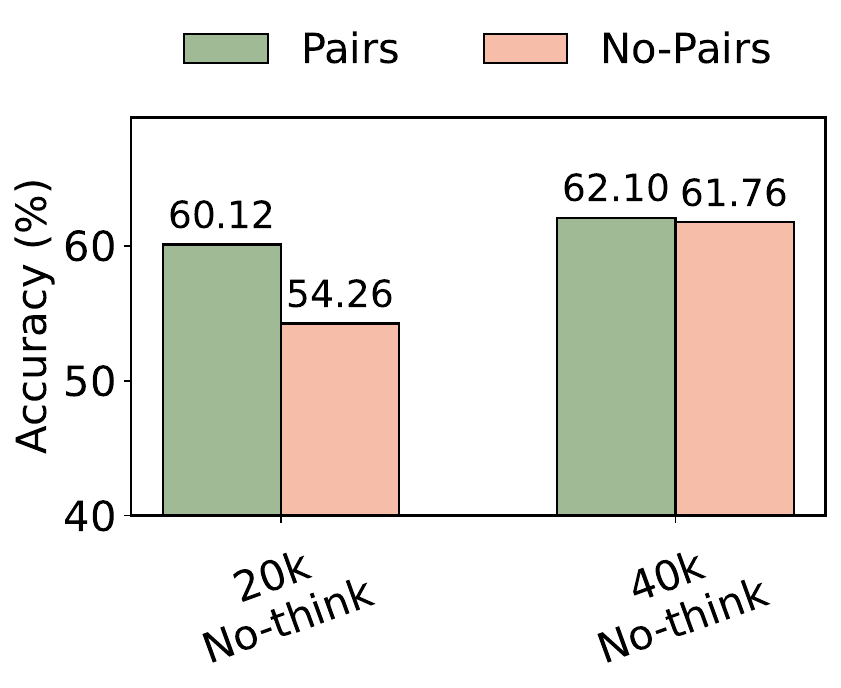}
    \caption{No-Think Acc.}
\end{subfigure}
\hfill
\begin{subfigure}[t]{0.24\linewidth}
    \centering
    \includegraphics[width=\linewidth]{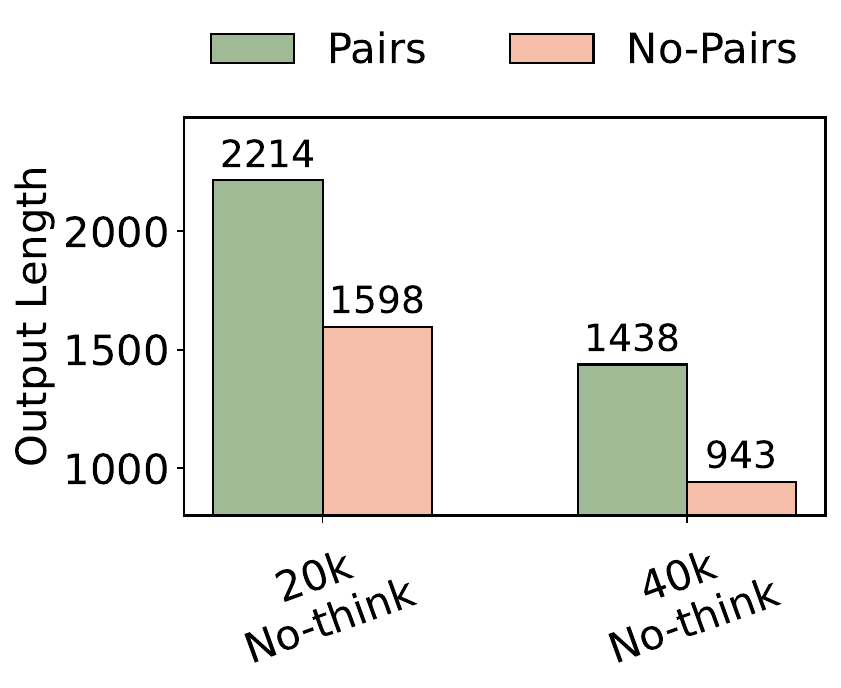}
    \caption{No-think Len.}
\end{subfigure}
\vspace{-6pt}
\caption{Comparison of accuracy and output lengths under think and no-think modes on MATH500. The bar charts illustrate model behavior under different training settings (pairs vs. no-pairs), highlighting how  no-pairs could reduce output length in no-think mode.}\label{fig:factor_pairs}
\end{figure*}

\section{Motivations: No-Think Mode is Still Thinking}

The implementation of hybrid thinking is typically achieved by adding control tokens 
(e.g., ``\texttt{\textbackslash no\textunderscore think}'', ``\texttt{\textbackslash think}'') into prompts and applying supervised fine-tuning (SFT), enabling models to seemingly decide whether to reason. To examine whether this provides true controllability, we evaluate Qwen3-8B~\citep{yang2025qwen3} on MATH500~\citep{lightman2023lets}, AIME24, and GPQA~\citep{rein2024gpqa}, with Qwen2.5-7B-Instruct serving as a pure no-think baseline. Evaluation results in \Cref{tab:mov_all} show that hybrid models consistently perform better in the think mode than in the no-think mode (e.g., Qwen3-8B achieves 63\% accuracy with an average length of 11,394 tokens in think mode, compared to 24\% and 4,062 tokens in no-think mode on AIME24), confirming that they learn to use control tokens to modulate output style and length.

However, comparison with Qwen2.5-7B-Instruct reveals that mode separation remains incomplete. As shown in \Cref{fig:mov_all} and \Cref{tab:mov_all}, Qwen3-8B  \textbf{ produces much longer no-think outputs} (e.g., 4,062 tokens on AIME24 task, compared to Qwen2.5-7B-Instruct's 1,729 tokens length), and \textbf{leaks reasoning-like behaviors} in its no-think responses (e.g., generating 646 ``\texttt{wait}'' tokens on AIME24 task, in contrast to Qwen2.5-7B-Instruct which operates strictly without reasoning traces).

As further illustrated in \Cref{fig:motivation}, when answering an AIME24 problem, Qwen3-8B, despite leaving the ``\texttt{<think>}'' reasoning block empty in no-think mode, inserted reasoning phrases such as ``\texttt{wait}'' when summarizing before the final answer. To confirm that such reflection markers indicate genuine reasoning rather than surface artifacts, and that the pattern extends beyond this single example, we further evaluate no-think outputs with an external LLM judge (GPT-OSS-20B) that classifies each response as \textit{no reasoning}, \textit{implicit reasoning} (the \texttt{<think></think>} block is empty but chain-of-thought reasoning or reflection is present in the visible output), or \textit{explicit reasoning} (chain-of-thought is properly contained within the \texttt{<think></think>} block). As shown in \Cref{tab:appendix_judge_validation}, across Qwen3-4B/8B/14B on MATH500, AIME24, and GPQA, the judge labels 92--100\% of no-think outputs as containing implicit reasoning, while think-mode outputs are labeled explicit reasoning 94--100\% of the time.

Together, these findings indicate that hybrid models can only partially suppress reasoning and fail to achieve a clean separation between think and no-think modes. Motivated by this limitation, our work focuses on systematically exploring training strategies and trade-offs to strengthen mode controllability, aiming to achieve finer and more reliable separation between think and no-think behaviors.
\section{What Factors Affect Mode Control in Hybrid Thinking Model Training}\label{sec:factors}

This section investigates training factors influencing the controllability of think and no-think mode within hybrid thinking. 
(1) \textit{Data Scale:} effective control requires sufficiently large data (e.g., 140k samples). 
(2) \textit{Pairing:} avoiding paired think/no-think samples—i.e., using answers from different questions—slightly improves no-think controllability compared to paired data.  
(3) \textit{Data Ratio:} while models easily learn the think mode, no-think is harder; moderately increasing its proportion enhances controllability. 
(4) \textit{Two-Phase Training:} training first on pure think data and then applying think-mode fusion yields stronger no-think control than mixed training.

All experiments use the OpenR1-Math~\citep{openr1} dataset, with no-think data from Numina-Math~\citep{numina_math_datasets} and think data generated by DeepSeek-R1~\citep{guo2025deepseek}.

\begin{figure*}[t]
\centering
\begin{subfigure}[t]{0.24\linewidth}
    \centering
    \includegraphics[width=\linewidth]{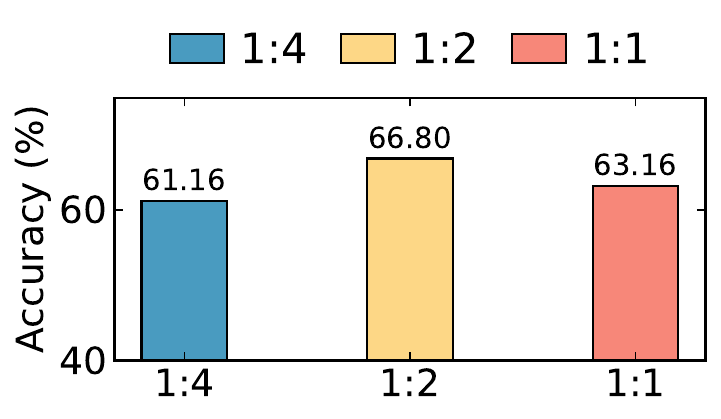}
    \caption{No-Think Acc.}
\end{subfigure}
\hfill
\begin{subfigure}[t]{0.24\linewidth}
    \centering
    \includegraphics[width=\linewidth]{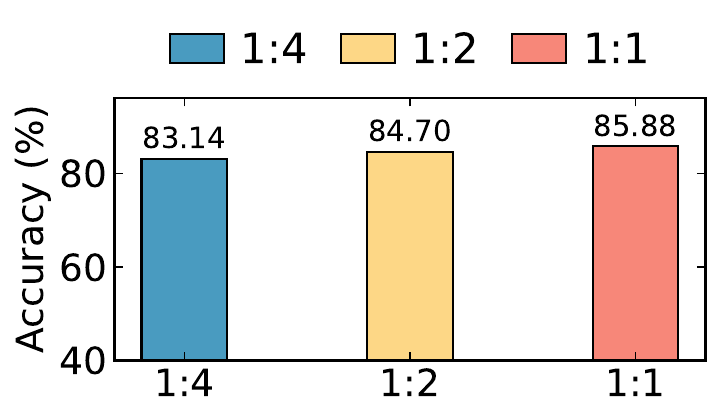}
    \caption{Think Acc.}
\end{subfigure}
\hfill
\begin{subfigure}[t]{0.24\linewidth}
    \centering
    \includegraphics[width=\linewidth]{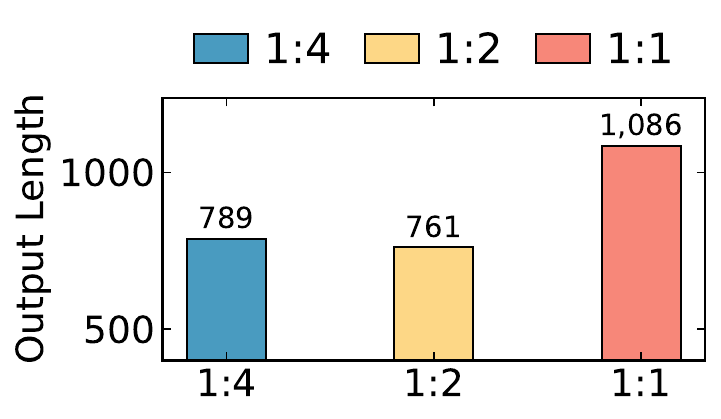}
    \caption{No-Think Len.}
    \label{fig:math500_ratio_acc}
\end{subfigure}
\hfill
\begin{subfigure}[t]{0.24\linewidth}
    \centering
    \includegraphics[width=\linewidth]{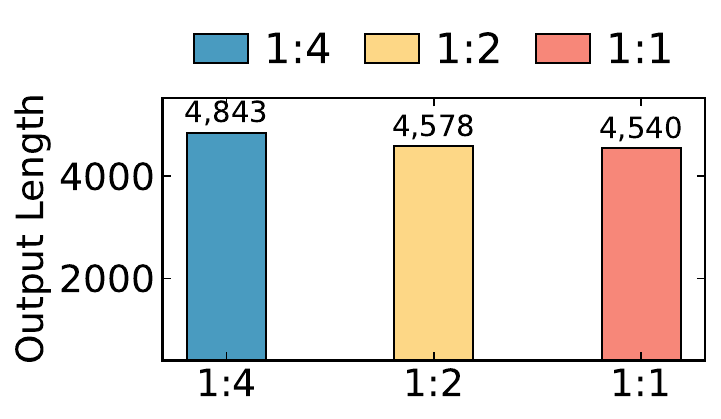}
    \caption{Think Len.}
    \label{fig:math500_ratio_len}
\end{subfigure}
\vspace{-6pt}
\caption{Comparison of output lengths under think and no-think modes on MATH500. 
The bar charts illustrate model behavior under different think-to-no-think data ratios, showing that increasing the proportion of no-think data reduces output length in the no-think mode.}
\label{fig:all_length}
\end{figure*}

\begin{figure*}[t]
\centering
\begin{subfigure}[t]{0.32\linewidth}
    \centering
    \includegraphics[width=\linewidth]{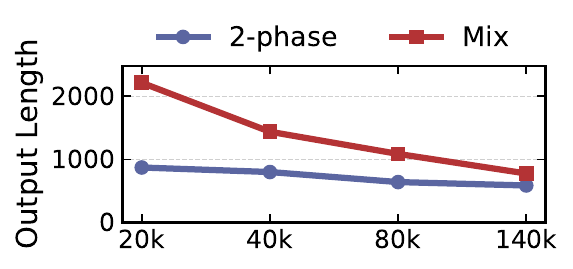}
    \caption{MATH500}
    \label{fig:Math500 — No-think Output Length}
\end{subfigure}
\hfill
\begin{subfigure}[t]{0.32\linewidth}
    \centering
    \includegraphics[width=\linewidth]{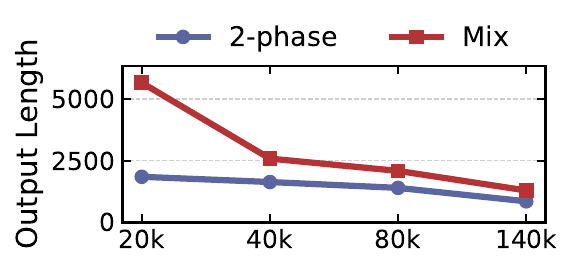}
    \caption{AIME24}
    \label{fig:AIME24 — No-think Output Length}
\end{subfigure}
\hfill
\begin{subfigure}[t]{0.32\linewidth}
    \centering
    \includegraphics[width=\linewidth]{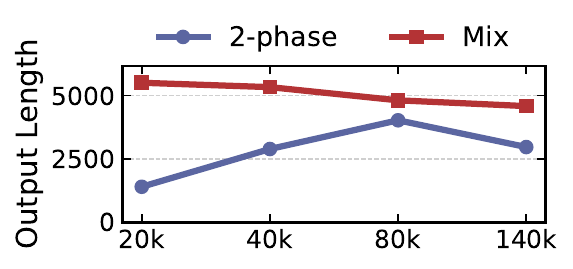}
    \caption{GPQA}
    \label{fig:GPQA — No-think Output Length}
\end{subfigure}
\vspace{-6pt}
\caption{Average output length of models trained with 2-phase and mix strategies across three benchmarks (MATH500, AIME24, and GPQA) under different data scales with no-think mode.}
\label{fig:No-think Accuracy}
\end{figure*}
\subsection{Large-Scale Data Facilitates Hybrid Thinking Mode Control}\label{sec:factor_data_scale}

We first explore the impact of \textit{data scale} on training hybrid thinking models. Specifically, we collect 20k, 40k, 80k, and 140k examples, ensuring that the ratio between think and no-think data is approximately 1:1. For each question, we include both a think-mode and a no-think-mode response. Comparative results for Qwen2.5-7B-Instruct are provided in \Cref{fig:line plot data scale}. For comprehensive and detailed experimental results, please refer to \Cref{tab:data_scale_math500,tab:data_scale_aime24,tab:data_scale_gpqa} for Qwen2.5-7B-Instruct, and \Cref{tab:data_scale_llama_math500,tab:data_scale_llama_aime24,tab:data_scale_llama_gpqa} for Llama3.1-8B-Instruct in Appendix~\ref{appendix:data_scale_results}.

We evaluate the model on MATH500, AIME24, and GPQA under both think and no-think modes. The results show that as the training data scale increases, the accuracy in the think mode steadily improves, while the output length remains roughly unchanged. In contrast, the accuracy in the no-think mode remains relatively stable, but the output length decreases significantly. At 140k scale, the no-think output length is the shortest, approaching that of the baseline. For example, on MATH500, after training with 140k data, the model’s no-think output length is reduced to 775.67, which is nearly comparable to the baseline length of 608.85. This indicates that training with 140k examples serves as an effective data scale for enabling the model to achieve hybrid thinking.
\label{sec: factor-pairs}

\subsection{Using Different Questions for Think and No-Think Improves Mode Control}

We further investigate the impact of \textit{With-Think vs No-Think Pairs} on hybrid thinking training. Using 20k and 40k training examples, we compare two settings: one where each question is paired with both a think-mode and a no-think-mode response, and another where the think and no-think responses are not paired to the same question.

As shown in \Cref{fig:factor_pairs}, consistent with our broader experimental results, we find that under the \textit{no-pairs} setting, models produce substantially shorter outputs in the no-think mode, while maintaining almost the same accuracy in both think and no-think modes. For example, at the 40k scale, the think accuracy of pairs and no-pairs is 85 and 84 respectively, and the no-think accuracy is 62\% and 61\%. However, the no-think output length of the no-pairs setting is only 942, significantly shorter than that of the pairs setting. This suggests that when constructing datasets, it is preferable to use non-paired settings—i.e., ensuring that think and no-think responses do not originate from the same question. For comprehensive experimental results, please refer to \Cref{tab:ht_control_math500,tab:ht_control_aime24,tab:ht_control_llama_math500,tab:ht_control_llama_aime24} for Qwen2.5-7B-Instruct and Llama3.1-8B-Instruct in Appendix~\ref{appendix:pairs_results}

\begin{table*}[t]
\centering
\caption{Comparison of hybrid, pure-think, and pure-no-think models trained from Qwen2.5-7B-Instruct on MATH500 and AIME24 under think and no-think modes. 
We report accuracy (Acc.), output length (Len.), and \#Wait count (i.e., occurrences of reflection words such as ``\texttt{wait}'', ``\texttt{hmm}'', ``\texttt{alternatively}''). 
Results show that in the no-think mode, hybrid models still produce much longer outputs than pure no-think models and generate frequent reflection words.}

\vspace{-8pt}
\label{tab:pure_models}
\begin{tabular}{ccccccc}
\toprule
\multirow{2}{*}{Model Type} 
& \multicolumn{3}{c}{Think} & \multicolumn{3}{c}{No-think} \\
\cmidrule(lr){2-4} \cmidrule(lr){5-7}
& Acc. & Len. & \#Wait & Acc. & Len. & \#Wait \\
\midrule
\multicolumn{7}{c}{MATH500} \\
\midrule
Pure think     & 86.66 & 4340.62 & 99149 & -- & -- & -- \\
Pure no-think  & -- & --  & -- & 53.10 & 430.44  & 0 \\
Hybrid think   & 86.12 & 4539.38 & 105655 & 77.42 & 2971.03 & \textbf{64622} \\
\midrule
\multicolumn{7}{c}{AIME24} \\
\midrule
Pure think     & 33.33 & 12324.62 & 14460 & -- & -- & -- \\
Pure no-think  & --  & --   & -- & 2.33  & 583.10   & 0 \\
Hybrid think   & 29.33 & 12615.44 & 15217 & 5 & 853.82  & \textbf{1} \\
\bottomrule
\end{tabular}
\end{table*}

\subsection{Moderately Higher No-Think Data Proportions Improve Mode Control}

Consistent with our earlier observations on data scaling in \Cref{sec:factor_data_scale}, we find that think-mode accuracy is relatively insensitive to scale, whereas no-think conciseness improves significantly. This suggests a practical lever for controllability: upweighting no-think data may strengthen the model's ability to remain concise in the no-think mode, thereby improving overall hybrid-thinking control.

To validate this hypothesis, we fine-tune the Qwen2.5-7B-Instruct and Llama3.1-8B-Instruct model with 80k training examples while controlling the ratio between think and no-think data, and then evaluate its hybrid thinking ability on MATH500 and AIME24. As shown in \Cref{fig:all_length}, with more comprehensive data in \Cref{tab:data_ratio_math500,tab:data_ratio_aime24,tab:data_ratio_llama_math500,tab:data_ratio_llama_aime24,tab:data_ratio_llama_gpqa,tab:data_ratio_llama_mmlu} provided in Appendix~\ref{appendix:data_ratio_results}, we find that appropriately increasing the no-think data proportion improves no-think mode controllability, as evidenced by shorter outputs, while leaving the accuracy and output length of the think mode, as well as the accuracy of the no-think mode, largely unaffected.

\subsection{Two-Phase Training: Training on Think Data Followed by Hybrid Thinking Training Enhances Mode Control}

In the previous sections, we collected both think and no-think data and directly shuffled them for fine-tuning, enabling the model to acquire hybrid thinking ability, which we called as mix training. In contrast, following the training strategy of models such as Qwen3-32B\citep{yang2025qwen3}, hybrid thinking can also be achieved through a \textit{two-phase training} process: first training the model on think-mode data to acquire reasoning ability, and then applying fine-tuning with both think and no-think data as a \textit{``Thinking Mode Fusion''} phase to integrate the no-think capabilities. 

To compare the effectiveness of two-phase training (think first, then think-mode fusion) against the random mixing strategy, we fine-tune the Qwen2.5-7B-Instruct and Llama3.1-8B-Instruct model with 80k examples, keeping the ratio of think and no-think pairs roughly 1:1 (an approximate proportion of 5:4). As shown in \Cref{fig:No-think Accuracy}, with the comprehensive experimental evidence detailed in \Cref{tab:training_strategies_math500,tab:training_strategies_aime24,tab:training_strategies_llama_math500,tab:training_strategies_llama_aime24} provided in Appendix~\ref{appendix:2-phase_results} , we observe that under different data scales, two-phase training consistently reduces output length in the no-think mode compared to mix training. For example, with 20k data, the no-think output length is reduced to 870 on MATH500 and 1847 on AIME24, compared to 2214 and 5654 under mix training.

The results demonstrate that two-phase training effectively reduces the output length in the no-think mode while preserving the model’s think-mode accuracy. This suggests that two-phase training mitigates the influence of think-mode data on no-think outputs, thereby improving no
think control.

We hypothesize this stems from \textit{interference} between reasoning-acquisition and mode-formatting objectives under mix training: think-mode data provides dense, high-magnitude gradient signals that dominate training and impede learning the concise no-think behavior, consistent with the degraded no-think controllability of hybrid vs.\ pure no-think models in \Cref{sec:pure-model-comparison}. We term this the \textit{Interference Hypothesis}. Two-phase training mitigates it by disentangling the objectives temporally: Phase 1 (pure think) solidifies reasoning capability in isolation; Phase 2 (thinking-mode fusion) then only needs to learn \textit{when} to reason rather than \textit{how} and \textit{when} simultaneously, freeing capacity for mode-routing and formatting. This explains why two-phase training yields shorter, more strictly formatted no-think outputs without sacrificing think-mode accuracy. This mirrors Qwen3's reasoning-first-then-fusion order~\citep{yang2025qwen3}; it remains a plausible interpretation, not one we have tested via dedicated ablation.

\section{How Can We Achieve True Hybrid Thinking?}

The previous sections mainly examined the training factors that influence hybrid thinking. In this section, we first compare hybrid thinking models with pure thinking and pure no-thinking models to analyze the trade-offs inherent in hybrid thinking. We also evaluate open-source models such as Qwen3 for further insights. Building on these analyses, we then propose a training recipe inspired by the findings of the previous section, with the goal of enhancing control over the no-think mode.

\textbf{Experimental Setups}. We sample 80K examples from the \textit{OpenR1-Math} default subset to construct three datasets: Pure think (only \texttt{think} responses), Pure no-Think (only \texttt{no\_think} responses), and Hybrid think (a $5\!:\!4$ mixture of \texttt{think} and \texttt{no\_think}). \textit{Qwen2.5-7B-Base} is fine-tuned on each dataset for 3 epochs with a learning rate of $1.0 \times 10^{-5}$ and a warm-up ratio of $0.1$, while all other optimization settings remain at their defaults.

\subsection{Limitations of Current Hybrid Thinking Training on Mode Switching}\label{sec:pure-model-comparison}

\begin{table*}[t]
\centering
\caption{Performance of Qwen3 models on MATH500 and AIME24 under think and no-think modes. 
We report accuracy (Acc.), output length (Len.), and \#Wait count. Results show that even open-source models still produce such reflection words in the no-think mode.}
\vspace{-8pt}
\label{tab:pure_models_qwen3}
\begin{tabular}{cccccccc}
\toprule
\multirow{2}{*}{Model} & \multirow{2}{*}{Mode} 
& \multicolumn{3}{c}{MATH500} & \multicolumn{3}{c}{AIME24} \\
\cmidrule(lr){3-5} \cmidrule(lr){6-8}
& & Acc. & Len. & \#Wait & Acc. & Len. & \#Wait \\
\midrule
\multirow{2}{*}{Qwen3-8B} 
 & Think     & 92.82 & 4384.35 & 83703 & 63.33 & 11394.54 & 12184 \\
 & No-think  & 82.90 & 958.31  & \textbf{646} & 24.00 & 4061.67  & \textbf{184} \\
\midrule
\multirow{2}{*}{Qwen3-4B} 
 & Think     & 92.02 & 4679.54 & 97998 & 61.67 & 11595.51 & 13754 \\
 & No-think  & 82.22 & 1004.13 & \textbf{611} & 20.67 & 4636.37  & \textbf{762} \\
\midrule
\multirow{2}{*}{Qwen3-14B} 
 & Think     & 93.96 & 4450.98 & 65320 & 67.67 & 11401.06 & 11091 \\
 & No-think  & 85.88 & 886.17  & \textbf{498} & 29.33 & 3497.86 & \textbf{295} \\
\bottomrule
\end{tabular}
\end{table*}

\begin{table*}[t]
\centering
\caption{Performance of different training settings on MATH500 and MMLU-STEM under think and no-think modes. We report accuracy (Acc.), output length (Len.), and \#Wait count (i.e., occurrences of reflection words such as ``\texttt{wait}'', ``\texttt{hmm}'', ``\texttt{alternatively}''). Our recipe maintains accuracy while substantially reducing no-think verbosity and reflective tokens.} 
\label{tab:two_phase}
\vspace{-8pt}
\setlength{\tabcolsep}{10pt}
\begin{tabular}{ccccccc}
\toprule
\multicolumn{7}{c}{MATH500} \\
\midrule
\multirow{2}{*}{Model} & \multicolumn{3}{c}{Think} & \multicolumn{3}{c}{No-think} \\
\cmidrule(lr){2-4} \cmidrule(lr){5-7}
& Acc. & Len. & \#Wait & Acc. & Len. & \#Wait \\
\midrule
Pure-think       & 86.66 & 4340.62 & 99149 & --    & --      & -- \\
Instruct         & --    & --      & --    & 75.42 & 640.91  & 0 \\
Original Recipe & 85.88 & 4539.53 & 116663 & 63.16 & 1085.99 & 5917 \\
Our Recipe  & 86.78 & 4481.74 & 103335 & 63.60 & \textbf{585.63} & \textbf{522} \\
\midrule
\multicolumn{7}{c}{MMLU-STEM} \\
\midrule
\multirow{2}{*}{Model} & \multicolumn{3}{c}{Think} & \multicolumn{3}{c}{No-think} \\
\cmidrule(lr){2-4} \cmidrule(lr){5-7}
& Acc. & Len. & \#Wait & Acc. & Len. & \#Wait \\
\midrule
Pure-think       & 84.97 & 2924.20 & 47382 & --    & --     & -- \\
Instruct         & --    & --      & --    & 66.25 & 427.01 & 18 \\
Original Recipe & 84.68 & 3129.53 & 52665 & 80.97 & 2014.49 & 24653 \\
Our Recipe  & 84.74 & 3141.14 & 50574 & 60.61 & \textbf{956.57} & \textbf{4017} \\
\bottomrule
\end{tabular}
\end{table*}

Since current hybrid thinking models lack direct pure-think or pure no-think counterparts, we train three variants from the same base model: a hybrid model (80k hybrid data), a pure-think model (80k think data), and a pure no-think model (80k no-think data). As shown in \Cref{tab:pure_models}, the hybrid model achieves nearly identical performance to the pure-think model in the think mode (e.g., MATH500 output length 4539 vs. 4340; AIME24 output length 12615 vs. 12324), suggesting that think-mode ability is unaffected. However, in the no-think mode, hybrid models achieve higher accuracy but generate substantially longer outputs than pure no-think models, indicating interference from think data that undermines controllability.  

We further examine open-source hybrid models such as Qwen3 in \Cref{tab:pure_models_qwen3}. Even with large-scale training data, these models still exhibit reasoning-related behaviors in the no-think mode. To quantify this phenomenon, we count the occurrences of reasoning-supportive tokens (e.g., ``\texttt{wait}'', ``\texttt{hmm}'', ``\texttt{alternatively}'') and observe frequent appearances across no-think outputs. For instance, all Qwen3 models (4B, 8B, and 14B) produce such tokens in the no-think setting. 

These findings underscore a fundamental limitation of current hybrid thinking models: they only partially suppress reasoning and thus fail to guarantee clean separation between modes.

\subsection{New Recipe for Improving No Think Control Based on Our Findings}\label{sec:new-recipe}

Although hybrid thinking models cannot yet achieve 100\% controllability over whether to engage in reasoning, the analysis in previous sections suggests practical strategies to improve their performance. The results on MATH500 and MMLU-STEM are reported in \Cref{tab:two_phase}. 

Here we use the pure-think model and the Qwen2.5-7B-Instruct model as baselines, representing the best achievable performance in the think and no-think modes, respectively. Inspired by the findings in~\Cref{sec:factors}, we take the 80k mixed training as the original recipe and compare it with our new 2-phase recipe trained on 140k data with pairs. Here, our recipe adopts paired data as a trade-off: as discussed in \Cref{sec: factor-pairs}, using unpaired think/no-think data can slightly improve controllability under the same data scale, but each instance is then used only once—either for think or for no-think—thus limiting the overall dataset size. We therefore prioritize enlarging the dataset at the cost of pairing, and the final results show that this scale-oriented recipe yields better overall performance. As shown, our recipe achieves nearly identical accuracy and output length to the original recipe in the think mode, but in the no-think mode it substantially reduces verbosity while maintaining accuracy—for example, on the MATH500 task, the average output length decreases from 1085 to 585 tokens, and the number of ``\texttt{wait}'' occurrences drops from 5917 to 522.

We additionally evaluate the recipe on the full MMLU benchmark (57 subjects) and TruthfulQA (\Cref{appendix:generalization}); reasoning leakage persists on general-knowledge and open-ended tasks beyond math, and our recipe's no-think accuracy remains close to the pure-instruct baseline while reflective-token counts drop by more than an order of magnitude relative to pure reasoning-tuned models.

We assess the recipe's stability and efficiency by repeating evaluation over 10 random seeds, reporting mean and standard deviation for output length and reflective-token count on MATH500 and AIME24, as well as wall-clock inference time and throughput on MATH500 (\Cref{appendix:efficiency}); the low variance confirms these reductions are stable, and no-think inference is consistently faster than think-mode inference across model families.

\begin{table*}[t]
\centering
\caption{The performance of proposed recipe across different models. We reported the performance of our recipe and the baseline pure think and pure no\_think models for Qwen2.5-7B, Llama3.1-8B, and Phi-4-14B. The results shows that our recipe consistently enhances performance across various model architectures (Qwen, Llama, and Phi) and scaling sizes (ranging from 7B to 14B), confirming its robustness and cross-architecture effectiveness. These results are from a different evaluation run than those in \Cref{tab:two_phase}.}
\vspace{-8pt}
\label{tab:recipes}
\begin{tabular}{cccccccc}
\toprule
\multirow{2}{*}{Model} & \multirow{2}{*}{Mode} 
& \multicolumn{3}{c}{MATH500} & \multicolumn{3}{c}{MMLU-STEM} \\
\cmidrule(lr){3-5} \cmidrule(lr){6-8}
& & Acc. & Len. & \#Wait & Acc. & Len. & \#Wait \\
\midrule
\multirow{4}{*}{Qwen2.5-7B} 
& Pure-think   & 86.06 & 4361.15 & 94997 & 85.79 & 2991.11 & 54345 \\
& Instruct  & 75.42 & 640.91  & 0 & 66.25 & 427.01 & 18 \\
& Our Recipe-think  & 86.50 & 4488.40 & 113183 & 85.82 & 3099.89 & 49397 \\
& Our Recipe-no-think  & 63.56 & \textbf{593.09} & \textbf{1535} & 59.44 & \textbf{774.42} & \textbf{2334} \\
\midrule
\multirow{4}{*}{Llama3.1-8B} 
& Pure-think     & 87.34 & 3814.88 & 96024 & 66.95 & 2089.36 & 43584 \\
& Instruct  & 49.50 & 2364.44 & 7 & 53.00 & 862.21 & 0 \\
& Our Recipe-think  & 80.46 & 5077.65 & 144079 & 78.94 & 3871.83 & 73016 \\
& Our Recipe-no-think  & 49.92 & \textbf{732.56} & \textbf{54} & 50.30 & \textbf{449.57} & \textbf{381} \\

\midrule
\multirow{4}{*}{Phi-4-14B} 
& Pure-think     & 92.00 & 3410.81 & 9502 & 70.57 & 2582.76 & 3213 \\
& Instruct  & 79.98 & 704.79 & 0 & 61.66 & 784.97 & 1 \\
& Our Recipe-think  & 92.48 & 3447.88 & 61706 & 93.63 & 2136.26 & 25489 \\
& Our Recipe-no-think  & 67.76 & \textbf{530.52} & \textbf{3} & 63.50 & \textbf{345.99} & \textbf{29} \\

\bottomrule
\end{tabular}
\end{table*}

\section{Related Works}

\textbf{LLM Reasoning.} Current approaches to enhancing reasoning in language models \citep{chen2025towards, plaat2024reasoning, sun2023survey} mainly fall into reinforcement learning (RL) \mbox{\citep{schulman2017proximal, jaech2024openai}} and supervised fine-tuning (SFT) \citep{yang2024qwen2}. In RL, DeepSeek \citep{guo2025deepseek, liu2024deepseek} achieved state-of-the-art performance with GRPO \citep{shao2024deepseekmath, yu2025dapo} and distilled reasoning traces to smaller models, inspiring replications such as Logic RL~\citep{xie2025logicrlunleashingllmreasoning} and SimpleRL-Zoo~\citep{zeng2025simplerlzooinvestigatingtamingzero}. RL and distillation effects on reasoning are also analyzed \citep{yeo2025demystifyinglongchainofthoughtreasoning, baek2025understandingdistilledreasoningmodels}. In SFT, curated reasoning traces are widely used, e.g., SkyThought-T1~\citep{sky_t1_2025} and Bespoke-Stratos-32B~\citep{bespoke_stratos}. Analyses show that structure of reasoning steps may matter more than content \citep{li2025llmseasilylearnreason}, while initial tokens play a key role \citep{ji2025first}. Data selection is also critical, with s1 \citep{muennighoff2025s1simpletesttimescaling} showing that small high-quality sets yield large gains. Other works probe complementary aspects of reasoning: CoT boundaries~\citep{chen2024unlocking}, distillation factors~\citep{chen2025unveiling}, RL dynamics~\citep{liu2025understanding}, and training-free multi-round thinking~\citep{tian2025thinktwiceenhancingllm}. Overall, RL and SFT provide complementary pathways to improve reasoning, highlighting the importance of training objectives, data structure, and sample quality.

\textbf{Efficient Reasoning.} Current reasoning models still face efficiency challenges, with many producing excessively long outputs \citep{bandyopadhyay2025thinking, li2025system}. Early efforts such as Kimi 1.5 \citep{team2025kimi} and Sky-Thought \citep{reduce_overthinking_2025} reduce verbosity by aligning long and short responses via preference optimization, while methods like TokenSkip \citep{xia2025tokenskip} and LightThinker \citep{zhang2025lightthinker} improve efficiency by removing redundant tokens or compressing intermediate thoughts. Beyond shortening reasoning, hybrid thinking \citep{sui2025stop} aims to control \textit{when} to reason by introducing control tokens (e.g., \texttt{\textbackslash think}, \nothink), a strategy now adopted by models such as Gemini, Qwen3, and DeepSeek V3.1 . However, current implementations still show imperfect mode separation, often leaking reasoning behavior into no-think outputs. Other works explore efficiency from different angles, such as early stopping with probing \citep{fu2024efficiently} or analyzing overthinking and underthinking behaviors \citep{wang2025thoughtsplaceunderthinkingo1like, sui2025stop, chen2024not}.

\textbf{Hybrid Thinking.} Hybrid thinking has recently emerged as a mechanism to balance efficiency and reasoning. It was first popularized by Gemini~\citep{comanici2025gemini} through an adjustable thinking budget that controls how much reasoning the model performs before answering, and later extended by Qwen3~\citep{yang2025qwen3} through explicit control tokens that switch between direct answering and reasoning across multiple model scales. GPT-oss~\citep{agarwal2025gpt} further explored varying degrees of reasoning depth, while DeepSeek V3.1~\citep{deepseekai2025v31} further extended the paradigm to improve agentic capabilities and efficiency. A concurrent line of work addresses this at the architecture level: Path-Lock Expert~\citep{wang2026pathlock} enforces mode separation via a fixed per-sequence routing decision through an existing interface token, rather than a learned router. Despite these advances, little work has systematically examined the training factors and trade-offs underlying hybrid thinking.

\section{Conclusion}
In this paper, we systematically analyzed hybrid thinking models, which balance efficiency and reasoning by switching between think and no-think modes. Our evaluation shows that while hybrid thinking enables partial controllability, current models still fail to fully separate the two modes. Through controlled experiments, we identified four key factors---data scale, paired vs.\ unpaired answers, data ratio, and two-phase training---with larger no-think proportions and two-phase training proving most effective. We conclude that hybrid thinking entails inherent trade-offs compared to pure-think or pure-no-think models, and recommend allocating moderately more no-think data and refining training strategies to further improve controllability of hybrid thinking models.

\clearpage

\section*{Limitations}
Our experiments are conducted primarily on earlier-generation models, including Qwen2.5 and Llama3.1, and do not yet cover the latest generation of models such as Qwen3. This choice is mainly due to practical considerations: at the time of our study, there's no officially released text-only instruction-tuned Qwen3-Instruct model available, and directly comparing against Qwen3-base models may not constitute a fair or controlled evaluation. We leave a systematic investigation of newer-generation models, including Qwen3 and beyond, to future work, which will help further validate and refine our conclusions.

\section*{Acknowledgments}
This work was supported in part by NSF awards 2117439 and 2112606. 

The authors used LLM-based assistants to help with LaTeX editing, drafting and polishing text, proofreading, and code implementation; all content was reviewed and verified by the authors.

\bibliography{references}

\clearpage
\appendix

\section*{Appendix} \label{sec:appendix}
\section{Additional Results on Factors Affect Hybrid Thinking}
\subsection{Results for Data Scales}\label{appendix:data_scale_results}

Comprehensive results for models trained at different scales (20k, 40k, 80k, and 140k samples) are provided in \Cref{tab:data_scale_math500,tab:data_scale_aime24,tab:data_scale_gpqa} for Qwen2.5-7B-Instruct and \Cref{tab:data_scale_llama_math500,tab:data_scale_llama_aime24,tab:data_scale_llama_gpqa} for Llama3.1-8B-Instruct. 

As shown in \Cref{tab:data_scale_math500,tab:data_scale_aime24,tab:data_scale_gpqa}, large-scale data is essential for effective control. While think-mode accuracy improves with scale (e.g., up to 86.58\% on MATH500), the most notable impact is the reduction in no-think length—decreasing from 2214.08 to 775.67 tokens on MATH500 as scale increases. This consistency across architectures confirms that 140k samples are vital for achieving mode separation and minimizing reasoning leakage.
\begin{table}[h!]
\centering
\caption{Performance of \textbf{Qwen2.5-7B-Instruct} with different data scales on \textbf{MATH500}. 
``Instruct'' refers to the original Qwen2.5-7B-Instruct model, while 20k/40k/80k/140k denote the number of training samples used. 
All training sets maintain a nearly 1:1 ratio of think to no-think data. 
We report accuracy (\%) and average output length under both modes for each model.}
\label{tab:data_scale_math500}
\vspace{-8pt}
\setlength{\tabcolsep}{3pt}
\begin{tabular}{ccccc}
\toprule
\multirow{2}{*}{Data Scale} & \multicolumn{2}{c}{Accuracy (\%)} & \multicolumn{2}{c}{Avg. Output Len.} \\
\cmidrule(lr){2-3} \cmidrule(lr){4-5}
 & Think & No-think & Think & No-think \\
\midrule
Instruct & --    & 74.96 & --    & 608.85 \\
20k      & 83.18 & 60.12 & 4704.91 & 2214.08 \\
40k      & 85.46 & 62.10 & 4589.04 & 1437.72 \\
80k      & 85.88 & 63.16 & 4539.53 & 1086.00 \\
140k     & 86.58 & 63.90 & 4442.49 & 775.67 \\
\bottomrule
\end{tabular}
\end{table}

\begin{table}[h!]
\centering
\caption{Performance of \textbf{Qwen2.5-7B-Instruct} with different data scales on \textbf{AIME24}. 
``Instruct'' refers to the original Qwen2.5-7B-Instruct model, while 20k/40k/80k/140k denote the number of training samples used. 
All training sets maintain a nearly 1:1 ratio of think to no-think data. 
We report accuracy (\%) and average output length under both modes for each model.}
\label{tab:data_scale_aime24}
\vspace{-8pt}
\setlength{\tabcolsep}{3pt}
\begin{tabular}{ccccc}
\toprule
\multirow{2}{*}{Data Scale} & \multicolumn{2}{c}{Accuracy (\%)} & \multicolumn{2}{c}{Avg. Output Len.} \\
\cmidrule(lr){2-3} \cmidrule(lr){4-5}
 & Think & No-think & Think & No-think \\
\midrule
Instruct & --    & 13.33 & --      & 1272.64 \\
20k      & 18.33 & 7.00  & 13397.91 &  5654.04 \\
40k      & 19.33 & 7.00  & 13586.36 & 2584.13 \\
80k      & 27.67 & 5.33  & 12799.13 & 2086.05 \\
140k     & 36.00 & 5.00  & 12507.64 & 1293.08 \\
\bottomrule
\end{tabular}
\end{table}

\begin{table}[h!]
\centering
\caption{Performance of \textbf{Qwen2.5-7B-Instruct} with different data scales on \textbf{GPQA}. 
``Instruct'' refers to the original Qwen2.5-7B-Instruct model, while 20k/40k/80k/140k denote the number of training samples used. 
All training sets maintain a nearly 1:1 ratio of think to no-think data. 
We report accuracy (\%) and average output length under both modes for each model.}
\label{tab:data_scale_gpqa}
\vspace{-8pt}
\setlength{\tabcolsep}{3pt}
\begin{tabular}{ccccc}
\toprule
\multirow{2}{*}{Data Scale} & \multicolumn{2}{c}{Accuracy (\%)} & \multicolumn{2}{c}{Avg. Output Len.} \\
\cmidrule(lr){2-3} \cmidrule(lr){4-5}
 & Think & No-think & Think & No-think \\
\midrule
Instruct & --    & 31.01 & --    & 977 \\
20k      & 40.00 & 39.50 & 9444.03 & 5503.32 \\
40k      & 40.61 & 38.99 & 9571.43 & 5330.81 \\
80k      & 40.25 & 38.89 & 9712.00 & 4811.29 \\
140k     & 41.01 & 34.95 & 9826.74 & 4583.18 \\
\bottomrule
\end{tabular}
\end{table}

\begin{table}[h!]
\centering
\caption{Performance of \textbf{Llama3.1-8B-Instruct} with different data scales on \textbf{MATH500}. 
``Instruct'' refers to the original Llama3.1-8B-Instruct model, while 20k/40k/80k/140k denote the number of training samples used. 
All training sets maintain a nearly 1:1 ratio (an approximate proportion of 5:4) of think to no-think data. 
We report accuracy (\%) and average output length under both modes for each model.}
\label{tab:data_scale_llama_math500}
\vspace{-8pt}
\setlength{\tabcolsep}{3pt}
\begin{tabular}{ccccc}
\toprule
\multirow{2}{*}{Data Scale} & \multicolumn{2}{c}{Accuracy (\%)} & \multicolumn{2}{c}{Avg. Output Len.} \\
\cmidrule(lr){2-3} \cmidrule(lr){4-5}
 & Think & No-think & Think & No-think \\
\midrule
Instruct & --    & 49.32 & --    & 2426.36 \\
20k      & 69.10 & 49.12 & 6008.2 & 1111.1 \\
40k      & 73.40 & 53.92 & 5685.3 & 1118.5 \\
80k      & 78.12 & 55.22 & 5263.7 & 899.7 \\
140k     & 82.00 & 57.72 & 4896.8 & 780.9 \\
\bottomrule
\end{tabular}
\end{table}

\begin{table}[h!]
\centering
\caption{Performance of \textbf{Llama3.1-8B-Instruct} with different data scales on \textbf{AIME24}. 
``Instruct'' refers to the original Llama3.1-8B-Instruct model, while 20k/40k/80k/140k denote the number of training samples used. 
All training sets maintain a nearly 1:1 ratio (an approximate proportion of 5:4) of think to no-think data. 
We report accuracy (\%) and average output length under both modes for each model.}
\label{tab:data_scale_llama_aime24}
\vspace{-8pt}
\setlength{\tabcolsep}{3pt}
\begin{tabular}{ccccc}
\toprule
\multirow{2}{*}{Data Scale} & \multicolumn{2}{c}{Accuracy (\%)} & \multicolumn{2}{c}{Avg. Output Len.} \\
\cmidrule(lr){2-3} \cmidrule(lr){4-5}
 & Think & No-think & Think & No-think \\
\midrule
Instruct & --    & 6.0   & --      & 7153.73 \\
20k      & 8.33  & 5.67  & 14182.8 & 2949.1 \\
40k      & 10.00 & 4.00  & 14088.2 & 2995.5 \\
80k      & 20.33 & 4.00  & 13253.2 & 2108.0 \\
140k     & 23.67 & 2.67  & 13141.8 & 2229.5 \\
\bottomrule
\end{tabular}
\end{table}

\begin{table}[h!]
\centering
\caption{Performance of \textbf{Llama3.1-8B-Instruct} with different data scales on \textbf{GPQA}. 
``Instruct'' refers to the original Llama3.1-8B-Instruct model, while 20k/40k/80k/140k denote the number of training samples used. 
All training sets maintain a nearly 1:1 ratio (an approximate proportion of 5:4) of think to no-think data. 
We report accuracy (\%) and average output length under both modes for each model.}
\label{tab:data_scale_llama_gpqa}
\vspace{-8pt}
\setlength{\tabcolsep}{3pt}
\begin{tabular}{ccccc}
\toprule
\multirow{2}{*}{Data Scale} & \multicolumn{2}{c}{Accuracy (\%)} & \multicolumn{2}{c}{Avg. Output Len.} \\
\cmidrule(lr){2-3} \cmidrule(lr){4-5}
 & Think & No-think & Think & No-think \\
\midrule
Instruct & --    & 29.29 & --     & 3190.01 \\
20k      & 36.41 & 32.53 & 9119.5 & 2814.4 \\
40k      & 36.52 & 30.35 & 10144.5 & 1957.3 \\
80k      & 37.58 & 31.16 & 10004.7 & 987.2 \\
140k     & 34.95 & 31.67 & 10190.1 & 3352.6 \\
\bottomrule
\end{tabular}
\end{table}

\newpage
\subsection{Results for Pairs and No-Pairs Data}\label{appendix:pairs_results}
Comprehensive results for models trained using data with pairs and without pairs are provided in \Cref{tab:ht_control_math500,tab:ht_control_aime24} for Qwen2.5-7B-Instruct and \Cref{tab:ht_control_llama_math500,tab:ht_control_llama_aime24} for Llama3.1-8B-Instruct. 

Results across scales and architectures reinforce that non-paired responses improve no-think controllability. As shown in \Cref{tab:ht_control_math500,tab:ht_control_aime24}, the no-pairs setting consistently yields shorter no-think outputs (e.g., 942.87 vs. 1437.72 tokens at 40k scale on MATH500) with comparable accuracy. This consistent pattern across Qwen2.5 and Llama3.1 suggests that decoupling responses helps models better reduces reasoning leakage.

\begin{table}[h]
\centering
\caption{Comparison between pairs and no-pairs on \textbf{MATH500} under think and no-think modes for  \textbf{Qwen2.5-7B-Instruct}. \textit{Pairs} means each question has both think and no-think responses; \textit{no-pairs} means responses come from different questions. We report accuracy (\%) and average output length. The no-pairs setting yields shorter no-think outputs across scales while maintaining comparable accuracy.}
\label{tab:ht_control_math500}
\vspace{-8pt}
\setlength{\tabcolsep}{3pt}
\begin{tabular}{cccccc}
\toprule
\multirow{2}{*}{Scale} & \multirow{2}{*}{Setting} & \multicolumn{2}{c}{Think} & \multicolumn{2}{c}{No-think} \\
\cmidrule(lr){3-4} \cmidrule(lr){5-6}
 & & Acc. & Len. & Acc. & Len. \\
\midrule
\multirow{2}{*}{20k} 
 & pairs     & 83.18 & 4704.91 & 60.12 & 2214.08 \\
 & no-pairs  & 83.44 & 4733.91 & 54.26 & \textbf{1597.64} \\
\midrule
\multirow{2}{*}{40k} 
 & pairs     & 85.46 & 4589.04 & 62.10 & 1437.72 \\
 & no-pairs  & 84.40 & 4560.66 & 61.76 & \textbf{942.87} \\
 \midrule
\multirow{2}{*}{80k} 
 & pairs     & 85.88 & 4539.53 & 63.16 & 1086.00 \\
 & no-pairs  & 86.6  & 4584.13 & 56.12 & \textbf{884.79} \\
\bottomrule
\end{tabular}
\end{table}

\begin{table}[h]
\centering
\caption{Comparison between pairs and no-pairs on \textbf{AIME24} under think and no-think modes for  \textbf{Qwen2.5-7B-Instruct}. \textit{Pairs} means each question has both think and no-think responses; \textit{no-pairs} means responses come from different questions. We report accuracy (\%) and average output length. No-pairs setting yields shorter no-think outputs across scales while maintaining accuracy.}
\label{tab:ht_control_aime24}
\vspace{-8pt}
\setlength{\tabcolsep}{3pt}
\begin{tabular}{cccccc}
\toprule
\multirow{2}{*}{Scale} & \multirow{2}{*}{Setting} & \multicolumn{2}{c}{Think} & \multicolumn{2}{c}{No-think} \\
\cmidrule(lr){3-4} \cmidrule(lr){5-6}
 & & Acc. & Len. & Acc. & Len. \\
\midrule
\multirow{2}{*}{20k} 
 & pairs     & 18.33 & 13397.91 & 7.00  & 5654.04 \\
 & no-pairs  & 23.00 & 12913.75 & 5.33  & \textbf{2560.70} \\
\midrule
\multirow{2}{*}{40k} 
 & pairs     & 19.33 & 13586.36 & 7.00  & 2584.13 \\
 & no-pairs  & 24.00 & 13308.04 & 7.67  & \textbf{1912.82} \\
\midrule
\multirow{2}{*}{80k} 
 & pairs     & 27.67 & 12799.13 & 5.33  & 2086.05 \\
 & no-pairs  & 32.67 & 12765.81 & 4.0   & \textbf{1146.767} \\
\bottomrule
\end{tabular}
\end{table}

\begin{table}[h!]
\centering
\caption{Comparison between pairs and no-pairs on \textbf{MATH500} under think and no-think modes for \textbf{Llama3.1-8B-Instruct}. \textit{Pairs} means each question has both think and no-think responses; \textit{no-pairs} means responses come from different questions. We report accuracy (\%) and average output length. The no-pairs setting yields shorter no-think outputs across scales while maintaining comparable accuracy.}
\label{tab:ht_control_llama_math500}
\vspace{-8pt}
\setlength{\tabcolsep}{3pt}
\begin{tabular}{cccccc}
\toprule
\multirow{2}{*}{Scale} & \multirow{2}{*}{Setting} & \multicolumn{2}{c}{Think} & \multicolumn{2}{c}{No-think} \\
\cmidrule(lr){3-4} \cmidrule(lr){5-6}
 & & Acc. & Len. & Acc. & Len. \\
\midrule
\multirow{2}{*}{20k} 
 & pairs     & 69.10 & 6008.17 & 49.12 & \textbf{1111.11} \\
 & no-pairs  & 69.50 & 6187.14 & 51.70 & 1159.26 \\
\midrule
\multirow{2}{*}{40k} 
 & pairs     & 73.40 & 5685.25 & 53.92 & 1118.49 \\
 & no-pairs  & 73.52 & 5679.92 & 54.34 & \textbf{1096.61} \\
\midrule
\multirow{2}{*}{80k} 
 & pairs     & 78.12 & 5263.71 & 55.22 & 899.72 \\
 & no-pairs  & 77.88 & 5353.67 & 56.32 & \textbf{864.91} \\
\bottomrule
\end{tabular}
\end{table}

\begin{table}[h!]
\centering
\caption{Comparison between pairs and no-pairs on \textbf{AIME24} under think and no-think modes for \textbf{Llama3.1-8B-Instruct}. \textit{Pairs} means each question has both think and no-think responses; \textit{no-pairs} means responses come from different questions. We report accuracy (\%) and average output length. The no-pairs setting yields shorter no-think outputs across scales while maintaining comparable accuracy.}
\label{tab:ht_control_llama_aime24}
\vspace{-8pt}
\setlength{\tabcolsep}{3pt}
\begin{tabular}{cccccc}
\toprule
\multirow{2}{*}{Scale} & \multirow{2}{*}{Setting} & \multicolumn{2}{c}{Think} & \multicolumn{2}{c}{No-think} \\
\cmidrule(lr){3-4} \cmidrule(lr){5-6}
 & & Acc. & Len. & Acc. & Len. \\
\midrule
\multirow{2}{*}{20k} 
 & pairs     & 8.33  & 14182.81 & 5.67  & 2949.06 \\
 & no-pairs  & 11.33 & 13880.53 & 3.33  & \textbf{2850.43} \\
\midrule
\multirow{2}{*}{40k} 
 & pairs     & 10.00 & 14088.24 & 4.00  & \textbf{2995.47} \\
 & no-pairs  & 13.33 & 13926.53 & 3.00  & 3203.56 \\
\midrule
\multirow{2}{*}{80k} 
 & pairs     & 20.33 & 13253.21 & 4.00  & 2108.03 \\
 & no-pairs  & 18.00 & 13437.94 & 2.67  & \textbf{1877.41} \\
\bottomrule
\end{tabular}
\end{table}

\newpage
\subsection{Results for Think-to-No-Think Data Ratios}\label{appendix:data_ratio_results}
Comprehensive results for models trained using data at different think-to-no-think ratios (1:4, 1:2, 1:1, 2:1) are provided in \Cref{tab:data_ratio_math500,tab:data_ratio_aime24} for Qwen2.5-7B-Instruct and \Cref{tab:data_ratio_llama_math500,tab:data_ratio_llama_aime24,tab:data_ratio_llama_gpqa,tab:data_ratio_llama_mmlu} for Llama3.1-8B-Instruct respectively. 

Experiments show that increasing no-think data proportions enhances control. Per \Cref{tab:data_ratio_math500,tab:data_ratio_aime24}, a 1:2 ratio reduces no-think length from 1086.00 to 761.30 tokens on MATH500 compared to 1:1, with stable think-mode accuracy. This confirms that upweighting no-think samples in the dataset is a practical lever for mode separation across various benchmarks and architectures.
\begin{table}[h!]
\centering
\caption{Effect of different think-to-no-think data ratios on \textbf{MATH500} under think and no-think modes for
\textbf{Qwen2.5-7B-Instruct}. Ratios denote the proportion of think to no-think samples, with the total training size fixed at 80k. Increasing the proportion of no-think data effectively reduces no-think output length while largely preserving accuracy in both modes. \textit{Note: The ``1:1'' ratio is an approximate proportion ($\approx 5\!:\!4$) due to the design for dataset construction.}}
\label{tab:data_ratio_math500}
\vspace{-8pt}
\setlength{\tabcolsep}{3pt}
\begin{tabular}{ccccc}
\toprule
\multirow{2}{*}{Think:No-think} & \multicolumn{2}{c}{Think} & \multicolumn{2}{c}{No-think} \\
\cmidrule(lr){2-3} \cmidrule(lr){4-5}
 & Acc. & Len. & Acc. & Len. \\
\midrule
1:4        & 83.14 & 4843.01 & 61.16 & 788.70 \\
1:2        & 84.70 & 4577.90 & 66.80 & \textbf{761.30} \\
1:1        & 85.88 & 4539.53 & 63.16 & 1086.00 \\
2:1        & 86.6  & 4452.42 & 62.48 & 1121.09 \\
\bottomrule
\end{tabular}
\end{table}

\begin{table}[h!]
\centering
\caption{Effect of different think-to-no-think data ratios on \textbf{AIME24} under think and no-think modes for
\textbf{Qwen2.5-7B-Instruct}. Ratios denote the proportion of think to no-think samples, with the total training size fixed at 80k. Increasing the proportion of no-think data effectively reduces no-think output length while largely preserving accuracy in both modes. \textit{Note: The ``1:1'' ratio is an approximate proportion ($\approx 5\!:\!4$) due to the design for dataset construction.}}
\label{tab:data_ratio_aime24}
\vspace{-8pt}
\setlength{\tabcolsep}{3pt}
\begin{tabular}{ccccc}
\toprule
\multirow{2}{*}{Think:No-think} & \multicolumn{2}{c}{Think} & \multicolumn{2}{c}{No-think} \\
\cmidrule(lr){2-3} \cmidrule(lr){4-5}
 & Acc. & Len. & Acc. & Len. \\
\midrule
1:4        & 20.56 & 13229.06 & 4.33  & \textbf{1453.21} \\
1:2        & 25.00 & 13233.21 & 3.00  & 1592.96 \\
1:1        & 27.67 & 12799.13 & 5.33  & 2086.05 \\
2:1        & 31.33 & 12524.66 & 3.33  & 2233.13 \\
\bottomrule
\end{tabular}
\end{table}

\begin{table}[h!]
\centering
\caption{Effect of different think-to-no-think data ratios on \textbf{MATH500} under think and no-think modes for \textbf{Llama3.1-8B-Instruct}. Ratios denote the proportion of think to no-think samples, with the total training size fixed at 80k. Increasing the proportion of no-think data effectively reduces no-think output length while largely preserving accuracy in both modes. \textit{Note: The ``1:1'' ratio is an approximate proportion ($\approx 5\!:\!4$) due to the design for dataset construction.}}
\label{tab:data_ratio_llama_math500}
\vspace{-8pt}
\setlength{\tabcolsep}{3pt}
\begin{tabular}{ccccc}
\toprule
\multirow{2}{*}{Think:No-think} & \multicolumn{2}{c}{Think} & \multicolumn{2}{c}{No-think} \\
\cmidrule(lr){2-3} \cmidrule(lr){4-5}
 & Acc. & Len. & Acc. & Len. \\
\midrule
1:4        & 69.56 & 6184.32 & 50.40 & \textbf{802.80} \\
1:2        & 74.32 & 5743.64 & 53.04 & 811.43 \\
1:1        & 78.12 & 5263.71 & 55.22 & 899.72 \\
2:1        & 77.96 & 5147.29 & 55.16 & 1025.80 \\
\bottomrule
\end{tabular}
\end{table}

\begin{table}[h!]
\centering
\caption{Effect of different think-to-no-think data ratios on \textbf{AIME24} under think and no-think modes for \textbf{Llama3.1-8B-Instruct}. Ratios denote the proportion of think to no-think samples, with the total training size fixed at 80k. Increasing the proportion of no-think data effectively reduces no-think output length while largely preserving accuracy in both modes. \textit{Note: The ``1:1'' ratio is an approximate proportion ($\approx 5\!:\!4$) due to the design for dataset construction.}}
\label{tab:data_ratio_llama_aime24}
\vspace{-8pt}
\setlength{\tabcolsep}{3pt}
\begin{tabular}{ccccc}
\toprule
\multirow{2}{*}{Think:No-think} & \multicolumn{2}{c}{Think} & \multicolumn{2}{c}{No-think} \\
\cmidrule(lr){2-3} \cmidrule(lr){4-5}
 & Acc. & Len. & Acc. & Len. \\
\midrule
1:4        & 5.33  & 14416.15 & 2.00  & \textbf{1234.35} \\
1:2        & 13.33 & 13767.91 & 3.33  & 1774.01 \\
1:1        & 20.33 & 13253.21 & 4.00  & 2108.03 \\
2:1        & 19.33 & 13328.44 & 4.67  & 2229.22 \\
\bottomrule
\end{tabular}
\end{table}

\begin{table}[h!]
\centering
\caption{Effect of different think-to-no-think data ratios on \textbf{GPQA} under think and no-think modes for \textbf{Llama3.1-8B-Instruct}. Ratios denote the proportion of think to no-think samples, with the total training size fixed at 80k. Increasing the proportion of no-think data effectively reduces no-think output length while largely preserving accuracy in both modes. \textit{Note: The ``1:1'' ratio is an approximate proportion ($\approx 5\!:\!4$) due to the design for dataset construction.}}
\label{tab:data_ratio_llama_gpqa}
\vspace{-8pt}
\setlength{\tabcolsep}{3pt}
\begin{tabular}{ccccc}
\toprule
\multirow{2}{*}{Think:No-think} & \multicolumn{2}{c}{Think} & \multicolumn{2}{c}{No-think} \\
\cmidrule(lr){2-3} \cmidrule(lr){4-5}
 & Acc. & Len. & Acc. & Len. \\
\midrule
1:4        & 32.73 & 10886.36 & 30.81 & 1470.06 \\
1:2        & 33.23 & 10493.00 & 31.11 & 1039.15 \\
1:1        & 37.58 & 10004.75 & 31.16 & \textbf{987.24} \\
2:1        & 38.48 & 9885.08  & 31.92 & 1359.05 \\
\bottomrule
\end{tabular}
\end{table}

\begin{table}[h]
\centering
\caption{Effect of different think-to-no-think data ratios on \textbf{MMLU-STEM} under think and no-think modes for \textbf{Llama3.1-8B-Instruct}. Ratios denote the proportion of think to no-think samples, with the total training size fixed at 80k. Increasing the proportion of no-think data effectively reduces no-think output length while largely preserving accuracy in both modes. \textit{Note: The ``1:1'' ratio is an approximate proportion ($\approx 5\!:\!4$) due to the design for dataset construction.}}
\label{tab:data_ratio_llama_mmlu}
\vspace{-8pt}
\setlength{\tabcolsep}{3pt}
\begin{tabular}{ccccc}
\toprule
\multirow{2}{*}{Think:No-think} & \multicolumn{2}{c}{Think} & \multicolumn{2}{c}{No-think} \\
\cmidrule(lr){2-3} \cmidrule(lr){4-5}
 & Acc. & Len. & Acc. & Len. \\
\midrule
1:4        & 75.45 & 4474.87 & 50.02 & 568.52 \\
1:2        & 77.74 & 4181.22 & 49.60 & \textbf{540.94} \\
1:1        & 79.16 & 3774.97 & 48.84 & 546.10 \\
2:1        & 79.80 & 3542.41 & 49.67 & 584.47 \\
\bottomrule
\end{tabular}
\end{table}

\newpage
\subsection{Results for 2-Phase and Mix Training}\label{appendix:2-phase_results}
Comprehensive results for models trained via 2-phase and mix training strategies are provided in \Cref{tab:training_strategies_math500,tab:training_strategies_aime24} for Qwen2.5-7B-Instruct and \Cref{tab:training_strategies_llama_math500,tab:training_strategies_llama_aime24,tab:training_strategies_llama_gpqa,tab:training_strategies_llama_mmlu} for Llama3.1-8B-Instruct, which confirm the advantage of two-phase training in improving no-think controllability. Compared to mix training, the two-phase approach consistently produces shorter no-think outputs across benchmarks—such as MATH500 (585.63 vs. 775.67 tokens at 140k scale)—without sacrificing think-mode accuracy. This confirms that sequential training on think and hybrid data effectively reduces reasoning leakage.

\begin{table}[h!]
\centering
\caption{Comparison of 2-phase and mix training strategies on \textbf{MATH500} under think and no-think modes for \textbf{Qwen2.5-7B-Instruct}. \textit{Scale} denotes the number of samples (e.g., 20k = 20,000). We report accuracy (\%) and average output length. The \textit{2-phase} strategy first trains on pure-think data before fine-tuning with mixed data, while the \textit{mix} strategy directly trains on mixed data. Across all scales, \textit{2-phase} strategy yields shorter no-think outputs compared to \textit{mix} training.}
\label{tab:training_strategies_math500}
\vspace{-8pt}
\setlength{\tabcolsep}{3pt}
\begin{tabular}{cccccc}
\toprule
\multirow{2}{*}{Scale} & \multirow{2}{*}{Setting} & \multicolumn{2}{c}{Think} & \multicolumn{2}{c}{No-think} \\
\cmidrule(lr){3-4} \cmidrule(lr){5-6}
 & & Acc. & Len. & Acc. & Len. \\
\midrule
\multirow{2}{*}{20k} 
 & 2-phase   & 83.38 & 4797.40 & 47.26 & \textbf{870.50} \\
 & mix       & 83.18 & 4704.91 & 60.12 & 2214.08 \\
\midrule
\multirow{2}{*}{40k} 
 & 2-phase   & 85.18 & 4670.88 & 43.52 & \textbf{798.87} \\
 & mix       & 85.46 & 4589.04 & 62.10 & 1437.72 \\
\midrule
\multirow{2}{*}{80k} 
 & 2-phase   & 85.86 & 4626.42 & 51.86 & \textbf{639.60} \\
 & mix       & 85.88 & 4539.53 & 63.16 & 1086.00 \\
\midrule
\multirow{2}{*}{140k} 
 & 2-phase   & 86.78 & 4481.74 & 63.60 & \textbf{585.63} \\
 & mix       & 86.58 & 4442.49 & 63.90 & 775.67 \\
\bottomrule
\end{tabular}
\end{table}

\begin{table}[h!]
\centering
\caption{Comparison of 2-phase and mix training strategies on \textbf{AIME24} under think and no-think modes for \textbf{Qwen2.5-7B-Instruct}. \textit{Scale} denotes the number of training samples (e.g., 20k = 20,000). We report accuracy (\%) and average output length. The \textit{2-phase} strategy first trains on pure-think data before fine-tuning with mixed data, while the \textit{mix} strategy directly trains on mixed data. Across all scales, \textit{2-phase} strategy yields shorter no-think outputs compared to \textit{mix} training.}
\label{tab:training_strategies_aime24}
\vspace{-8pt}
\setlength{\tabcolsep}{3pt}
\begin{tabular}{cccccc}
\toprule
\multirow{2}{*}{Scale} & \multirow{2}{*}{Setting} & \multicolumn{2}{c}{Think} & \multicolumn{2}{c}{No-think} \\
\cmidrule(lr){3-4} \cmidrule(lr){5-6}
 & & Acc. & Len. & Acc. & Len. \\
\midrule
\multirow{2}{*}{20k} 
 & 2-phase   & 20.00 & 13620.71 & 4.33  & \textbf{1847.35} \\
 & mix       & 18.33 & 13397.91 & 7.00  & 5654.04 \\
\midrule
\multirow{2}{*}{40k} 
 & 2-phase   & 22.67 & 13562.19 & 1.67  & \textbf{1635.34} \\
 & mix       & 19.33 & 13586.36 & 7.00  & 2584.13 \\
\midrule
\multirow{2}{*}{80k} 
 & 2-phase   & 30.00 & 12962.41 & 3.00  & \textbf{1398.48} \\
 & mix       & 27.67 & 12799.13 & 5.33  & 2086.05 \\
\midrule
\multirow{2}{*}{140k} 
 & 2-phase   & 32.67 & 12271.01 & 5.00  & \textbf{853.82} \\
 & mix       & 36.00 & 12507.64 & 5.00  & 1293.08 \\
\bottomrule
\end{tabular}
\vspace{3em}
\end{table}

\begin{table}[h!]
\centering
\caption{Comparison of 2-phase and mix training strategies on \textbf{MATH500} under think and no-think modes for \textbf{Llama3.1-8B-Instruct}. Here, \textit{Scale} denotes the number of training samples (e.g., 20k means 20,000 samples). We report both accuracy (\%) and average output length. The \textit{2-phase} strategy first trains on pure-think data before fine-tuning with mixed think and no-think data, while the \textit{mix} strategy directly trains on think and no-think mixed data. Notably, across all scales, \textit{2-phase} strategy consistently yields shorter output lengths under the no-think mode compared to \textit{mix} training.}
\label{tab:training_strategies_llama_math500}
\vspace{-8pt}
\setlength{\tabcolsep}{3pt}
\begin{tabular}{cccccc}
\toprule
\multirow{2}{*}{Scale} & \multirow{2}{*}{Setting} & \multicolumn{2}{c}{Think} & \multicolumn{2}{c}{No-think} \\
\cmidrule(lr){3-4} \cmidrule(lr){5-6}
 & & Acc. & Len. & Acc. & Len. \\
\midrule
\multirow{2}{*}{20k} 
 & 2-phase   & 68.56 & 6300.03 & 47.18 & 1180.40 \\
 & mix       & 69.10 & 6008.17 & 49.12 & \textbf{1111.11} \\
\midrule
\multirow{2}{*}{40k} 
 & 2-phase   & 72.52 & 6033.99 & 48.50 & 1170.26 \\
 & mix       & 73.40 & 5685.25 & 53.92 & \textbf{1118.49} \\
\midrule
\multirow{2}{*}{80k} 
 & 2-phase   & 76.90 & 5475.04 & 50.00 & \textbf{870.21} \\
 & mix       & 78.12 & 5263.71 & 55.22 & 899.72 \\
\midrule
\multirow{2}{*}{140k} 
 & 2-phase   & 80.66 & 5059.14 & 50.76 & \textbf{731.98} \\
 & mix       & 82.00 & 4896.76 & 57.72 & 780.87 \\
\bottomrule
\end{tabular}
\vspace{4em}
\end{table}

\begin{table}[h!]
\centering
\caption{Comparison of 2-phase and mix training strategies on \textbf{AIME24} under think and no-think modes for \textbf{Llama3.1-8B-Instruct}. Here, \textit{Scale} denotes the number of training samples (e.g., 20k means 20,000 samples). We report both accuracy (\%) and average output length. The \textit{2-phase} strategy first trains on pure-think data before fine-tuning with mixed think and no-think data, while the \textit{mix} strategy directly trains on think and no-think mixed data. Notably, across all scales, \textit{2-phase} strategy consistently yields shorter output lengths under the no-think mode compared to \textit{mix} training.}
\label{tab:training_strategies_llama_aime24}
\vspace{-8pt}
\setlength{\tabcolsep}{3pt}
\begin{tabular}{cccccc}
\toprule
\multirow{2}{*}{Scale} & \multirow{2}{*}{Setting} & \multicolumn{2}{c}{Think} & \multicolumn{2}{c}{No-think} \\
\cmidrule(lr){3-4} \cmidrule(lr){5-6}
 & & Acc. & Len. & Acc. & Len. \\
\midrule
\multirow{2}{*}{20k} 
 & 2-phase   & 7.00  & 13737.77 & 4.33  & \textbf{2735.59} \\
 & mix       & 8.33  & 14182.81 & 5.67  & 2949.06 \\
\midrule
\multirow{2}{*}{40k} 
 & 2-phase   & 9.67  & 14034.31 & 4.33  & \textbf{2692.68} \\
 & mix       & 10.00 & 14088.24 & 4.00  & 2995.47 \\
\midrule
\multirow{2}{*}{80k} 
 & 2-phase   & 16.00 & 13690.04 & 5.00  & \textbf{1776.30} \\
 & mix       & 20.33 & 13253.21 & 4.00  & 2108.03 \\
\midrule
\multirow{2}{*}{140k} 
 & 2-phase   & 20.33 & 12837.85 & 1.67  & \textbf{1429.10} \\
 & mix       & 23.67 & 13141.84 & 2.67  & 2229.50 \\
\bottomrule
\end{tabular}
\end{table}

\begin{table}[h!]
\centering
\caption{Comparison of 2-phase and mix training strategies on \textbf{GPQA} under think and no-think modes for \textbf{Llama3.1-8B-Instruct}. Here, \textit{Scale} denotes the number of training samples (e.g., 20k means 20,000 samples). We report both accuracy (\%) and average output length. The \textit{2-phase} strategy first trains on pure-think data before fine-tuning with mixed think and no-think data, while the \textit{mix} strategy directly trains on think and no-think mixed data. Notably, across all scales, \textit{2-phase} strategy consistently yields shorter output lengths under the no-think mode compared to \textit{mix} training.}
\label{tab:training_strategies_llama_gpqa}
\vspace{-8pt}
\setlength{\tabcolsep}{3pt}
\begin{tabular}{cccccc}
\toprule
\multirow{2}{*}{Scale} & \multirow{2}{*}{Setting} & \multicolumn{2}{c}{Think} & \multicolumn{2}{c}{No-think} \\
\cmidrule(lr){3-4} \cmidrule(lr){5-6}
 & & Acc. & Len. & Acc. & Len. \\
\midrule
\multirow{2}{*}{20k} 
 & 2-phase   & 36.31 & 10182.25 & 26.67 & \textbf{966.86} \\
 & mix       & 36.41 & 9119.54  & 32.53 & 2814.42 \\
\midrule
\multirow{2}{*}{40k} 
 & 2-phase   & 35.15 & 10626.91 & 29.60 & \textbf{975.13} \\
 & mix       & 36.52 & 10144.51 & 30.35 & 1957.28 \\
\midrule
\multirow{2}{*}{80k} 
 & 2-phase   & 37.22 & 10100.31 & 27.98 & \textbf{705.01} \\
 & mix       & 37.58 & 10004.75 & 31.16 & 987.24 \\
\midrule
\multirow{2}{*}{140k} 
 & 2-phase   & 35.10 & 10077.06 & 28.48 & \textbf{745.44} \\
 & mix       & 34.95 & 10190.07 & 31.67 & 3352.59 \\
\bottomrule
\end{tabular}
\vspace{4em}
\end{table}

\begin{table}[h!]
\centering
\caption{Comparison of 2-phase and mix training strategies on \textbf{MMLU-STEM} under think and no-think modes for \textbf{Llama3.1-8B-Instruct}. Here, \textit{Scale} denotes the number of training samples (e.g., 20k means 20,000 samples). We report both accuracy (\%) and average output length. The \textit{2-phase} strategy first trains on pure-think data before fine-tuning with mixed think and no-think data, while the \textit{mix} strategy directly trains on think and no-think mixed data. Notably, across all scales, \textit{2-phase} strategy consistently yields shorter output lengths under the no-think mode compared to \textit{mix} training.}
\label{tab:training_strategies_llama_mmlu}
\vspace{-8pt}
\setlength{\tabcolsep}{3pt}
\begin{tabular}{cccccc}
\toprule
\multirow{2}{*}{Scale} & \multirow{2}{*}{Setting} & \multicolumn{2}{c}{Think} & \multicolumn{2}{c}{No-think} \\
\cmidrule(lr){3-4} \cmidrule(lr){5-6}
 & & Acc. & Len. & Acc. & Len. \\
\midrule
\multirow{2}{*}{20k} 
 & 2-phase   & 77.26 & 4056.91 & 53.38 & \textbf{572.71} \\
 & mix       & 76.82 & 3632.27 & 59.88 & 1127.23 \\
\midrule
\multirow{2}{*}{40k} 
 & 2-phase   & 79.10 & 3973.26 & 53.98 & \textbf{592.56} \\
 & mix       & 78.50 & 3866.03 & 55.22 & 765.63 \\
\midrule
\multirow{2}{*}{80k} 
 & 2-phase   & 79.10 & 3980.86 & 46.02 & \textbf{448.93} \\
 & mix       & 79.16 & 3774.97 & 48.84 & 546.10 \\
\midrule
\multirow{2}{*}{140k} 
 & 2-phase   & 78.50 & 3822.84 & 48.40 & \textbf{461.48} \\
 & mix       & 78.43 & 3782.77 & 52.52 & 849.36 \\
\bottomrule
\end{tabular}
\end{table}

\newpage
\section{Additional Results on Training Strategies}
\label{appendix:two_phase_mix}

\Cref{tab:datasets_math500,tab:datasets_aime24,tab:datasets_gpqa} report results of two-phase and mix training across scales on Math500, AIME24, and GPQA-Diamond. Two-phase consistently reduces no-think output length while preserving think-mode accuracy, showing its advantage in suppressing unnecessary reasoning traces without harming task performance.

\begin{table}[h!]
\centering
\caption{Performance of \textbf{Qwen2.5-7B-Instruct} under different training settings on \textbf{MATH500}. 
We report accuracy (Acc.) and output length (Len.) for both think and no-think modes.}
\label{tab:datasets_math500}
\vspace{-8pt}
\setlength{\tabcolsep}{3pt}
\begin{tabular}{lccccc}
\toprule
\multirow{2}{*}{Setting} & \multicolumn{2}{c}{Think} & \multicolumn{2}{c}{No-think} \\
\cmidrule(lr){2-3} \cmidrule(lr){4-5}
 & Acc. & Len. & Acc. & Len. \\
\midrule
Baseline          & --    & --      & 74.96 & 608.85 \\
Demo (w/o pairs)  & 37.08 & 650.35  & 31.68 & 390.63 \\
20k-2phase        & 83.38 & 4797.40 & 47.26 & 870.50 \\
20k-mix           & 83.18 & 4704.91 & 60.12 & 2214.08 \\
40k-2phase        & 85.18 & 4670.88 & 43.52 & 798.87 \\
40k-mix           & 85.46 & 4589.04 & 62.10 & 1437.72 \\
\bottomrule
\end{tabular}
\end{table}

\begin{table}[h!]
\centering
\caption{Performance of \textbf{Qwen2.5-7B-Instruct} under different training settings on \textbf{AIME24}. 
We report accuracy (Acc.) and output length (Len.) for both think mode and no-think mode.}
\label{tab:datasets_aime24}
\vspace{-8pt}
\setlength{\tabcolsep}{3pt}
\begin{tabular}{lccccc}
\toprule
\multirow{2}{*}{Setting} & \multicolumn{2}{c}{Think} & \multicolumn{2}{c}{No-think} \\
\cmidrule(lr){2-3} \cmidrule(lr){4-5}
 & Acc. & Len. & Acc. & Len. \\
\midrule
20k-2phase        & 20.00 & 13620.71 & 4.33  & 1847.35 \\
20k-mix           & 18.33 & 13397.91 & 7.00  & 5654.04 \\
40k-2phase        & 22.67 & 13562.19 & 1.67  & 1635.34 \\
40k-mix           & 19.33 & 13586.36 & 7.00  & 2584.13 \\
\bottomrule
\end{tabular}
\end{table}

\begin{table}[h!]
\centering
\caption{Performance of \textbf{Qwen2.5-7B-Instruct} under different training settings on \textbf{GPQA-Diamond}. 
We report accuracy (Acc.) and output length (Len.) for both think mode and no-think mode.}
\label{tab:datasets_gpqa}
\vspace{-8pt}
\setlength{\tabcolsep}{3pt}
\begin{tabular}{lccccc}
\toprule
\multirow{2}{*}{Setting} & \multicolumn{2}{c}{Think} & \multicolumn{2}{c}{No-think} \\
\cmidrule(lr){2-3} \cmidrule(lr){4-5}
 & Acc. & Len. & Acc. & Len. \\
\midrule
20k-2phase        & 41.16 & 9872.89 & 26.72 & 1404.46 \\
20k-mix           & 40.00 & 9444.03 & 39.50 & 5503.32 \\
40k-2phase        & 41.62 & 9903.88 & 32.12 & 2892.41 \\
40k-mix           & 40.61 & 9571.43 & 38.99 & 5330.81 \\
\bottomrule
\end{tabular}
\end{table}

\newpage
\section{System Prompt Ablation}
\label{appendix:system_prompt}

Table~\ref{tab:two_phase_sys_prompt} compares Qwen 2-phase training with and without system prompts on Math500, AIME24, and GPQA-Diamond. Results show that while system prompts slightly reduce output length, they also lead to degraded accuracy in no-think mode.
\begin{table}[h!]
\centering
\caption{Comparison of \textbf{Qwen} 2-phase training with and without system prompt on \textbf{MATH500}, \textbf{AIME24}, and \textbf{GPQA-Diamond}.}
\label{tab:two_phase_sys_prompt}
\vspace{-8pt}
\setlength{\tabcolsep}{5pt}
\begin{tabular}{lccccc}
\toprule
\multirow{2}{*}{Model} & \multicolumn{2}{c}{Think} & \multicolumn{2}{c}{No-think} \\
\cmidrule(lr){2-3} \cmidrule(lr){4-5}
 & Acc. & Len. & Acc. & Len. \\
\midrule
\multicolumn{5}{c}{Math500} \\
\midrule
w/o Sys. Prompt & 86.1 & 4539 & 77.4 & 2971 \\
w/ Sys. Prompt & 84.6 & 4383 & 76.6 & 2840 \\
\midrule
\multicolumn{5}{c}{AIME24} \\
\midrule
w/o Sys. Prompt & 29.3 & 12615 & 23.3 & 5744 \\
w/ Sys. Prompt & 28.7 & 11799 & 16.0 & 6269 \\
\midrule
\multicolumn{5}{c}{GPQA-Diamond} \\
\midrule
w/o Sys. Prompt & 40.6 & 10283 & 35.9 & 8378 \\
w/ Sys. Prompt & 39.0 & 9931 & 37.6 & 7784 \\
\bottomrule
\end{tabular}
\end{table}

\section{Generalization Beyond Mathematical Reasoning}\label{appendix:generalization}
This section reports the full results supporting the generalization evaluation in \Cref{sec:new-recipe}: accuracy, output length, and reflective-token count on the full MMLU benchmark (57 subjects) and TruthfulQA (\Cref{tab:gen_qwen_mmlu,tab:gen_qwen_truthfulqa,tab:gen_llama_mmlu,tab:gen_llama_truthfulqa,tab:gen_phi_mmlu,tab:gen_phi_truthfulqa}). Throughout this section, ``Instruct'' and ``Reasoning'' denote the pure-instruct and pure-reasoning-tuned baselines for each model family, and ``Our Recipe'' denotes our proposed hybrid-thinking training recipe in both modes.

\begin{table}[h!]
\centering
\caption{Generalization beyond math for \textbf{Qwen} on \textbf{full MMLU} (57 subjects; not the MMLU-STEM subset used elsewhere in this paper): accuracy, output length, and reflective-token count.}
\label{tab:gen_qwen_mmlu}
\vspace{-8pt}
\small
\setlength{\tabcolsep}{3pt}
\begin{tabular*}{\columnwidth}{@{\extracolsep{\fill}}lccc@{}}
\toprule
Setting & Acc. & Len. & \#Refl. \\
\midrule
Instruct & 46.17\% & 267.60 & 13 \\
Reasoning & 75.24\% & 3723.06 & 516808 \\
Our Recipe (think) & 76.93\% & 2822.63 & 274499 \\
Our Recipe (no-think) & 57.06\% & 1547.63 & 9073 \\
\bottomrule
\end{tabular*}
\end{table}

\begin{table}[h!]
\centering
\caption{Generalization beyond math for \textbf{Qwen} on \textbf{TruthfulQA}: accuracy, output length, and reflective-token count.}
\label{tab:gen_qwen_truthfulqa}
\vspace{-8pt}
\small
\setlength{\tabcolsep}{3pt}
\begin{tabular*}{\columnwidth}{@{\extracolsep{\fill}}lccc@{}}
\toprule
Setting & Acc. & Len. & \#Refl. \\
\midrule
Instruct & 69.0\% & 17.2 & 0 \\
Reasoning & 42.8\% & 22283.3 & 32220 \\
Our Recipe (think) & 61.5\% & 6474.3 & 437 \\
Our Recipe (no-think) & 56.5\% & 735.6 & 80 \\
\bottomrule
\end{tabular*}
\end{table}

\begin{table}[h!]
\centering
\caption{Generalization beyond math for \textbf{Llama} on \textbf{full MMLU} (57 subjects; not the MMLU-STEM subset used elsewhere in this paper): accuracy, output length, and reflective-token count.}
\label{tab:gen_llama_mmlu}
\vspace{-8pt}
\small
\setlength{\tabcolsep}{3pt}
\begin{tabular*}{\columnwidth}{@{\extracolsep{\fill}}lccc@{}}
\toprule
Setting & Acc. & Len. & \#Refl. \\
\midrule
Instruct & 64.32\% & 578.48 & 13 \\
Reasoning & 61.32\% & 1869.46 & 242607 \\
Our Recipe (think) & 70.89\% & 3748.36 & 491453 \\
Our Recipe (no-think) & 52.12\% & 531.77 & 14858 \\
\bottomrule
\end{tabular*}
\end{table}

\begin{table}[h!]
\centering
\caption{Generalization beyond math for \textbf{Llama} on \textbf{TruthfulQA}: accuracy, output length, and reflective-token count.}
\label{tab:gen_llama_truthfulqa}
\vspace{-8pt}
\small
\setlength{\tabcolsep}{3pt}
\begin{tabular*}{\columnwidth}{@{\extracolsep{\fill}}lccc@{}}
\toprule
Setting & Acc. & Len. & \#Refl. \\
\midrule
Instruct & 49.5\% & 9.2 & 0 \\
Reasoning & 54.2\% & 6210.8 & 8276 \\
Our Recipe (think) & 58.8\% & 3564.1 & 3916 \\
Our Recipe (no-think) & 41.0\% & 1256.9 & 0 \\
\bottomrule
\end{tabular*}
\end{table}

\begin{table}[h!]
\centering
\caption{Generalization beyond math for \textbf{Phi} on \textbf{full MMLU} (57 subjects; not the MMLU-STEM subset used elsewhere in this paper): accuracy, output length, and reflective-token count.}
\label{tab:gen_phi_mmlu}
\vspace{-8pt}
\small
\setlength{\tabcolsep}{3pt}
\begin{tabular*}{\columnwidth}{@{\extracolsep{\fill}}lccc@{}}
\toprule
Setting & Acc. & Len. & \#Refl. \\
\midrule
Instruct & 70.47\% & 933.33 & 21 \\
Reasoning & 76.05\% & 7485.83 & 32478 \\
Our Recipe (think) & 87.44\% & 1881.09 & 158644 \\
Our Recipe (no-think) & 65.36\% & 365.22 & 2906 \\
\bottomrule
\end{tabular*}
\end{table}

\begin{table}[h!]
\centering
\caption{Generalization beyond math for \textbf{Phi} on \textbf{TruthfulQA}: accuracy, output length, and reflective-token count.}
\label{tab:gen_phi_truthfulqa}
\vspace{-8pt}
\small
\setlength{\tabcolsep}{3pt}
\begin{tabular*}{\columnwidth}{@{\extracolsep{\fill}}lccc@{}}
\toprule
Setting & Acc. & Len. & \#Refl. \\
\midrule
Instruct & 75.0\% & 25.5 & 0 \\
Reasoning & 74.8\% & 257.6 & 0 \\
Our Recipe (think) & 70.5\% & 4335.6 & 6447 \\
Our Recipe (no-think) & 67.0\% & 159.0 & 0 \\
\bottomrule
\end{tabular*}
\end{table}

\section{Stability and Inference Efficiency}\label{appendix:efficiency}
This section reports the full results supporting the stability and efficiency evaluation in \Cref{sec:new-recipe}: mean and standard deviation over 10 random seeds for output length, reflective-token count, wall-clock inference time, and throughput. Throughout this section, ``Instruct'' and ``Reasoning'' denote the pure-instruct and pure-reasoning-tuned baselines for each model family, and ``Our Recipe'' denotes our proposed hybrid-thinking training recipe in both modes.

\begin{table}[h!]
\centering
\caption{Statistical rigor for \textbf{Qwen} on \textbf{MATH500}: mean $\pm$ std over 10 independent random seeds for reflective-token count (\#Refl.) and total output length (Len.).}
\label{tab:var_qwen_math500}
\vspace{-8pt}
\scriptsize
\setlength{\tabcolsep}{1.5pt}
\begin{tabular*}{\columnwidth}{@{\extracolsep{\fill}}lcc@{}}
\toprule
Setting & \#Refl. & Len. \\
\midrule
Instruct & $0.00 \pm 0.00$ & $329.00 \pm 105.86$ \\
Reasoning & $1346.00 \pm 282.37$ & $4251.20 \pm 420.17$ \\
Our Recipe (think) & $1335.50 \pm 130.81$ & $4396.16 \pm 330.28$ \\
Our Recipe (no-think) & $0.80 \pm 1.75$ & $529.41 \pm 119.65$ \\
\bottomrule
\end{tabular*}
\end{table}

\begin{table}[h!]
\centering
\caption{Statistical rigor for \textbf{Qwen} on \textbf{AIME24}: mean $\pm$ std over 10 independent random seeds for reflective-token count (\#Refl.) and total output length (Len.).}
\label{tab:var_qwen_aime24}
\vspace{-8pt}
\scriptsize
\setlength{\tabcolsep}{1.5pt}
\begin{tabular*}{\columnwidth}{@{\extracolsep{\fill}}lcc@{}}
\toprule
Setting & \#Refl. & Len. \\
\midrule
Instruct & $0.00 \pm 0.00$ & $927.37 \pm 1792.48$ \\
Reasoning & $255.30 \pm 56.50$ & $13013.27 \pm 2491.25$ \\
Our Recipe (think) & $223.90 \pm 26.91$ & $11962.17 \pm 2716.82$ \\
Our Recipe (no-think) & $0.00 \pm 0.00$ & $1222.47 \pm 1630.07$ \\
\bottomrule
\end{tabular*}
\end{table}

\begin{table}[h!]
\centering
\caption{Statistical rigor for \textbf{Llama} on \textbf{MATH500}: mean $\pm$ std over 10 independent random seeds for reflective-token count (\#Refl.) and total output length (Len.).}
\label{tab:var_llama_math500}
\vspace{-8pt}
\scriptsize
\setlength{\tabcolsep}{1.5pt}
\begin{tabular*}{\columnwidth}{@{\extracolsep{\fill}}lcc@{}}
\toprule
Setting & \#Refl. & Len. \\
\midrule
Instruct & $0.00 \pm 0.00$ & $1554.07 \pm 880.38$ \\
Reasoning & $1971.80 \pm 246.43$ & $4231.61 \pm 424.62$ \\
Our Recipe (think) & $2578.50 \pm 853.49$ & $5580.90 \pm 452.84$ \\
Our Recipe (no-think) & $0.90 \pm 1.91$ & $728.17 \pm 233.68$ \\
\bottomrule
\end{tabular*}
\end{table}

\begin{table}[h!]
\centering
\caption{Statistical rigor for \textbf{Llama} on \textbf{AIME24}: mean $\pm$ std over 10 independent random seeds for reflective-token count (\#Refl.) and total output length (Len.).}
\label{tab:var_llama_aime24}
\vspace{-8pt}
\scriptsize
\setlength{\tabcolsep}{1.5pt}
\begin{tabular*}{\columnwidth}{@{\extracolsep{\fill}}lcc@{}}
\toprule
Setting & \#Refl. & Len. \\
\midrule
Instruct & $0.00 \pm 0.00$ & $4635.40 \pm 4255.10$ \\
Reasoning & $352.80 \pm 125.31$ & $11714.47 \pm 3547.74$ \\
Our Recipe (think) & $438.30 \pm 436.76$ & $14688.33 \pm 1594.26$ \\
Our Recipe (no-think) & $0.40 \pm 1.26$ & $1105.03 \pm 600.98$ \\
\bottomrule
\end{tabular*}
\end{table}

\begin{table}[h!]
\centering
\caption{Statistical rigor for \textbf{Phi} on \textbf{MATH500}: mean $\pm$ std over 10 independent random seeds for reflective-token count (\#Refl.) and total output length (Len.).}
\label{tab:var_phi_math500}
\vspace{-8pt}
\scriptsize
\setlength{\tabcolsep}{1.5pt}
\begin{tabular*}{\columnwidth}{@{\extracolsep{\fill}}lcc@{}}
\toprule
Setting & \#Refl. & Len. \\
\midrule
Instruct & $0.00 \pm 0.00$ & $656.06 \pm 58.05$ \\
Reasoning & $127.90 \pm 36.85$ & $3010.61 \pm 1336.74$ \\
Our Recipe (think) & $1072.70 \pm 121.07$ & $3493.02 \pm 334.76$ \\
Our Recipe (no-think) & $0.00 \pm 0.00$ & $654.51 \pm 243.35$ \\
\bottomrule
\end{tabular*}
\end{table}

\begin{table}[h!]
\centering
\caption{Statistical rigor for \textbf{Phi} on \textbf{AIME24}: mean $\pm$ std over 10 independent random seeds for reflective-token count (\#Refl.) and total output length (Len.).}
\label{tab:var_phi_aime24}
\vspace{-8pt}
\scriptsize
\setlength{\tabcolsep}{1.5pt}
\begin{tabular*}{\columnwidth}{@{\extracolsep{\fill}}lcc@{}}
\toprule
Setting & \#Refl. & Len. \\
\midrule
Instruct & $0.00 \pm 0.00$ & $1229.37 \pm 429.75$ \\
Reasoning & $28.30 \pm 14.07$ & $6825.90 \pm 2727.26$ \\
Our Recipe (think) & $186.50 \pm 53.14$ & $10215.10 \pm 1813.10$ \\
Our Recipe (no-think) & $0.00 \pm 0.00$ & $645.17 \pm 190.61$ \\
\bottomrule
\end{tabular*}
\end{table}

\begin{table}[h!]
\centering
\caption{Statistical rigor for \textbf{Qwen} on \textbf{MATH500}: mean $\pm$ std over 10 independent random seeds for wall-clock inference time (s/instance) and throughput (tokens/s). Latency profiling was conducted on MATH500 only.}
\label{tab:lat_qwen}
\vspace{-8pt}
\scriptsize
\setlength{\tabcolsep}{1.5pt}
\begin{tabular*}{\columnwidth}{@{\extracolsep{\fill}}lcc@{}}
\toprule
Setting & Time (s) & Tok/s \\
\midrule
Instruct & $14.30 \pm 34.67$ & $314.66 \pm 73.98$ \\
Reasoning & $116.60 \pm 3.68$ & $333.54 \pm 55.39$ \\
Our Recipe (think) & $111.25 \pm 14.24$ & $321.88 \pm 51.86$ \\
Our Recipe (no-think) & $16.86 \pm 33.80$ & $330.18 \pm 67.98$ \\
\bottomrule
\end{tabular*}
\end{table}

\begin{table}[h!]
\centering
\caption{Statistical rigor for \textbf{Llama} on \textbf{MATH500}: mean $\pm$ std over 10 independent random seeds for wall-clock inference time (s/instance) and throughput (tokens/s). Latency profiling was conducted on MATH500 only.}
\label{tab:lat_llama}
\vspace{-8pt}
\scriptsize
\setlength{\tabcolsep}{1.5pt}
\begin{tabular*}{\columnwidth}{@{\extracolsep{\fill}}lcc@{}}
\toprule
Setting & Time (s) & Tok/s \\
\midrule
Instruct & $70.27 \pm 61.74$ & $238.27 \pm 88.20$ \\
Reasoning & $114.16 \pm 24.28$ & $306.27 \pm 55.07$ \\
Our Recipe (think) & $130.46 \pm 4.39$ & $339.03 \pm 46.82$ \\
Our Recipe (no-think) & $11.86 \pm 7.88$ & $297.78 \pm 49.75$ \\
\bottomrule
\end{tabular*}
\end{table}

\begin{table}[h!]
\centering
\caption{Statistical rigor for \textbf{Phi} on \textbf{MATH500}: mean $\pm$ std over 10 independent random seeds for wall-clock inference time (s/instance) and throughput (tokens/s). Latency profiling was conducted on MATH500 only.}
\label{tab:lat_phi}
\vspace{-8pt}
\scriptsize
\setlength{\tabcolsep}{1.5pt}
\begin{tabular*}{\columnwidth}{@{\extracolsep{\fill}}lcc@{}}
\toprule
Setting & Time (s) & Tok/s \\
\midrule
Instruct & $20.54 \pm 9.87$ & $187.35 \pm 21.99$ \\
Reasoning & $143.61 \pm 66.79$ & $153.76 \pm 36.66$ \\
Our Recipe (think) & $190.49 \pm 16.70$ & $160.36 \pm 21.07$ \\
Our Recipe (no-think) & $9.89 \pm 2.39$ & $195.37 \pm 23.20$ \\
\bottomrule
\end{tabular*}
\end{table}

\section{Broader Impact and Future Applications}

\paragraph{Agents.} In persistent agent workflows, most steps are routine (parsing a tool output, filling a template) rather than genuinely deliberative, and unnecessary reasoning compounds cost over long horizons rather than staying local~\citep{cuadron2025danger}; efficient serving of such long reasoning-augmented programs is itself an active target~\citep{fu2024efficiently}. A reliable no-think mode would let a system spend tokens only where deliberation is needed, which matters for persistent multi-agent workspaces where many calls are routine bookkeeping~\citep{wang2026agentteamworkzone}, for skill libraries that could declare a reasoning mode at call time~\citep{yang2026agentskills}, for evaluation over long, controllable-difficulty horizons~\citep{yang2026agentcebench}, and for keeping long-term memory records compact rather than padded with abandoned reasoning~\citep{long2026memtrace}.

\paragraph{Inference Serving and Efficiency.} On the serving side, most inference-time efficiency work targets the cost of producing each token -- through shorter reasoning traces~\citep{xu2025chain,lee2025well}, step-compressed reasoning~\citep{zhang2025lightthinker}, or broader surveys of efficient reasoning~\citep{sui2025stop} -- or through serving infrastructure such as paged attention~\citep{kwon2023efficient}. A reliable no-think mode instead reduces how many tokens are generated in the first place, and composes with these orthogonal axes: unbiased KV-cache compression via Fourier-domain windowing~\citep{li2025faedkv}, quantization strategies applied selectively to different parts of the KV cache~\citep{hariri2026quantize}, lossless weight compression via entropy coding~\citep{zhang2025dfloat11}, and efficient architecture and inference techniques explored for specialized language models in other domains such as protein sequence models~\citep{wang2026proteinlm}.

\paragraph{Model Architecture.} A separate line of work intervenes at the level of reasoning strategy or context handling rather than the token stream: surveys of reasoning strategies in LLMs~\citep{li2025system,bandyopadhyay2025thinking} and inference-time reasoning guidance~\citep{sui2025meta} motivate architecture-level control as a complementary axis. Training-free triggers can elicit an intermediate reasoning budget between full chain-of-thought and a direct answer~\citep{yang2026midthink}, context windows can be extended without additional tuning~\citep{jin2024longlm}, and mode separation itself can be enforced architecturally, with a fixed per-sequence routing decision rather than a learned router~\citep{wang2026pathlock}.

\paragraph{Post-Training and Evaluation.} The no-think behavior studied here is a property of post-training rather than architecture alone~\citep{rafailov2023direct}. Post-training effects can interact across domains in order-sensitive, asymmetric ways, directly relevant to reasoning-focused reinforcement learning~\citep{yang2026domains,yu2025dapo,shao2024deepseekmath}: mode separation should be re-measured whenever the training mixture changes rather than assumed to transfer. Inference-time scaling raises its own evaluation questions, from regime and reproducibility standards~\citep{hariri2026testtime,muennighoff2025s1simpletesttimescaling} to ranking reasoning models under test-time compute~\citep{hariri2026ranking}, and to guiding a small model with a larger one at inference time rather than retraining~\citep{yang2025speculative,leviathan2023fast}.

\end{document}